\documentclass[5p,twocolumn]{elsarticle}
\usepackage{latexsym}
\usepackage{graphicx}
\usepackage{amsfonts,amssymb,amsmath}
\usepackage{hyperref}
\usepackage[utf8]{inputenc}
\usepackage[T1]{fontenc}
\usepackage{tabularx}
\usepackage{appendix}
\usepackage{rotating}
\usepackage{booktabs}

\usepackage[table]{xcolor} 
\definecolor{lightblue}{rgb}{0.68, 0.85, 0.9}

\def\BibTeX{{\rm B\kern-.05em{\sc i\kern-.025em b}\kern-.08em
    T\kern-.1667em\lower.7ex\hbox{E}\kern-.125emX}}

\DeclareUnicodeCharacter{0301}{\'{e}}

\usepackage{enumitem}

\usepackage{microtype}

\usepackage{changepage}

\usepackage{graphicx}

\usepackage{caption}
\usepackage{subcaption}
\usepackage{soul}

\usepackage[linesnumbered,ruled]{algorithm2e}
\usepackage{algorithmicx,algpseudocode}
\usepackage{ifthen}
\usepackage{amsfonts}

\usepackage{hyperref,xcolor}

\usepackage{color, colortbl}
\definecolor{Gray}{gray}{0.9}
\definecolor{Gray2}{gray}{0.85}
\definecolor{Gray3}{rgb}{0.94,0.94,0.94}
\definecolor{Blue}{rgb}{0.8,0.83,0.95}
\definecolor{LightBlue}{rgb}{0.929,0.964,0.992}
\definecolor{LightGreen}{rgb}{0.898,0.996,0.803} 

\usepackage{amsmath,amssymb}

\usepackage{dsfont}

\newlength{\Oldarrayrulewidth}

\usepackage{multirow}
\usepackage{dsfont}

\usepackage{comment}

\definecolor{Gray}{gray}{0.70}
\definecolor{Gray2}{gray}{0.90}
\definecolor{LightCyan}{rgb}{0.88,1,1}

\newcolumntype{b}{>{\columncolor{Gray}}c}
\newcolumntype{a}{>{\columncolor{Gray2}}c}
\newcolumntype{d}{>{\columncolor{LightCyan}}c}

\newtheorem{definition}{Definition}[section]

\makeatletter
\def\ps@pprintTitle{%
  \let\@oddhead\@empty
  \let\@evenhead\@empty
  \def\@oddfoot{\reset@font\hfil}
  \let\@evenfoot\@oddfoot}
\makeatother

\begin{document}

\begin{frontmatter}

\title{Recent Advances in Traffic Accident Analysis and Prediction: A Comprehensive Review of Machine Learning Techniques\tnoteref{tnote1}}

\tnotetext[tnote1]{This paper is currently under review.}

\author[gmu]{Noushin Behboudi}
\ead{nbehboud@gmu.edu}

\author[osu]{Sobhan Moosavi}
\ead{moosavi.3@osu.edu}

\author[osu]{Rajiv Ramnath}
\ead{ramnath@cse.ohio-state.edu}

\address[gmu]{College of Science, George Mason University, Fairfax, Virginia}
\address[osu]{Department of Computer Science and Engineering, Ohio State University, Columbus, Ohio}

\begin{abstract}
Traffic accidents pose a severe global public health issue, leading to 1.19 million fatalities annually, with the greatest impact on individuals aged 5 to 29 years old. This paper addresses the critical need for advanced predictive methods in road safety by conducting a comprehensive review of recent advancements in applying machine learning (ML) techniques to traffic accident analysis and prediction. It examines 191 studies from the last five years, focusing on predicting accident risk, frequency, severity, duration, as well as general statistical analysis of accident data. To our knowledge, this study is the first to provide such a comprehensive review, covering the state-of-the-art across a wide range of domains related to accident analysis and prediction. The review highlights the effectiveness of integrating diverse data sources and advanced ML techniques to improve prediction accuracy and handle the complexities of traffic data. By mapping the current landscape and identifying gaps in the literature, this study aims to guide future research towards significantly reducing traffic-related deaths and injuries by 2030, aligning with the World Health Organization (WHO) targets. 
\end{abstract}

\begin{keyword}
Traffic Accident Prediction, Machine Learning, Deep Learning, Statistical Modeling
\end{keyword}

\end{frontmatter}

\section{Introduction}
\label{sec:introduction}
\begin{figure}[ht!]
    \centering
    \includegraphics[width=\columnwidth]{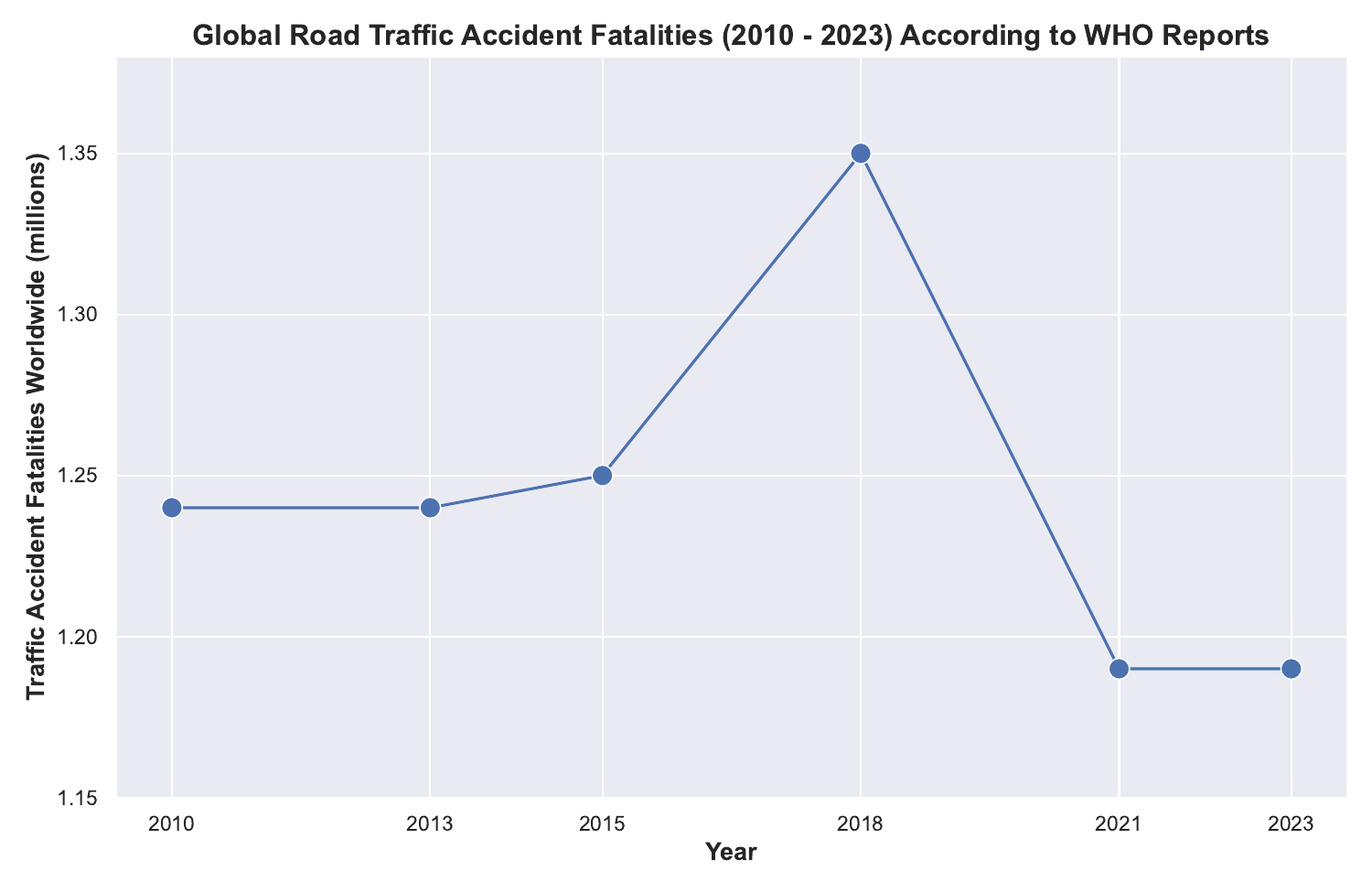}
    \caption{An overview of global road traffic accident fatalities (2010 - 2023) according to the WHO reports (\cite{WHO2010,WHO2013,WHO2015,WHO2018,WHO2021,WHO2023})}.
    \label{fig:who_reports}
\end{figure}

Traffic accidents remain a severe global public health crisis, as emphasized by the WHO in its 2023 Global Status Report on Road Safety~\cite{WHO2023}. Despite a slight reduction in overall fatalities to 1.19 million annually (see Figure~\ref{fig:who_reports}), traffic-related incidents are still the leading cause of death for individuals aged 5 to 29 years old, underscoring the ongoing gravity of the issue. The recent WHO report highlights that vulnerable road users, particularly in low- and middle-income countries, bear a substantial burden, with pedestrians, cyclists, and motorcyclists accounting for a significant proportion of traffic deaths. The report also notes that while over half of UN Member States have seen a decrease in road traffic deaths, significant disparities persist, especially in regions that lack adequate road safety measures and infrastructure. These challenges necessitate innovative solutions, specifically through the application of ML techniques. ML offers powerful tools for analyzing large datasets and identifying patterns that can predict and prevent road accidents. This paper explores the application of ML in enhancing road safety, contributing to the global effort to reduce traffic-related deaths and injuries by 2030, aligning with the WHO targets. 

Despite the critical importance of enhancing road safety through advanced predictive analytics, a thorough review of recent developments in machine-learning-based techniques for traffic accident analysis and prediction is notably lacking in the current literature. To the best of the authors' knowledge, no recent publication comprehensively reviews studies in the domain, especially those from the past five years, which are crucial for defining the state-of-the-art and charting directions for future research. This gap is significant given the rapid advancements in machine learning and the increasing availability of diverse and large-scale datasets relevant to traffic management.

Existing review studies cover a variety of topics, including accident risk, frequency, and severity prediction using machine learning techniques. These studies provide details on the methodologies used, data sources, and potential avenues for future research. 
Among the notable recent review studies, Silva et al. \cite{silva2020machine} provide a systematic review of ML in road safety, focusing on crash frequency, severity classification, and combined models. They discuss the advantages of ML in managing complex data to improve prediction accuracy, which is critical for developing effective traffic safety interventions. Gutierrez-Osorio and Pedraza \cite{gutierrez2020modern} review modern data sources and ML techniques for analyzing and forecasting road accidents. They emphasize the effectiveness of combining multiple analytical models to enhance prediction accuracy. Their analysis highlights the importance of integrating diverse data sources, including government records, onboard equipment data, and social media inputs, to deepen insights into traffic patterns and accident predictions. In another study, Sarkar and Maiti \cite{sarkar2020machine} comprehensively review the application of ML techniques in occupational accident analysis through a structured four-phase process: bibliometric search, descriptive analysis, scientometric analysis, and citation network analysis. Meanwhile, Angarita-Zapata et al. \cite{angarita2021bibliometric} present a bibliometric analysis complemented by benchmarks of ML and Automated Machine Learning (AutoML)\footnote{List of all acronyms in the paper are shared in Appendix~\ref{sec:appendix}.} tools like AutoGluon and CatBoost, tested in Colombian cities. Their findings suggest that general-purpose AutoML tools efficiently handle model selection, emphasizing their robust performance in crash severity prediction tasks. Wen et al. \cite{wen2021applications} review the use of ML in modeling traffic crash severity, noting challenges such as data imbalance and model transferability. They advocate for advanced ML methods, such as graph convolutional networks, to capture spatial and temporal correlations more effectively. Similarly, Santos et al. \cite{santos2022literature} analyze over 25 ML algorithms for predicting injury severity, with Random Forest and Support Vector Machines (SVM) showing promising results. They call for exploring diverse ML techniques to enhance predictive reliability and road safety management. Chand et al. \cite{chand2021road} review data sources, analysis techniques, and algorithms used in road traffic accident analysis, emphasizing key factors contributing to these incidents. Their paper provides an overview of the data sources available for road accident studies and evaluates various techniques for predicting the occurrence and severity of accidents. It also highlights operational challenges in traffic management and assesses the effectiveness of road safety measures. Ali et al. \cite{ali2024advances} conduct a comprehensive survey of 213 papers spanning from 1997 to 2023, exploring advances in ML-based crash prediction models specifically in the areas of crash occurrence, frequency, and severity prediction. Their systematic review identifies and addresses challenges such as unobserved heterogeneity and spatiotemporal dependencies, which are crucial in the context of road safety. 

Our study builds on this existing body of work to bring in the latest advancements and extend the range of the application areas discussed. In particular, this study fills a notable gap in the existing literature with respect to the machine learning techniques employed in traffic accident analysis and prediction. We address this gap by examining the research published in the last five years. We examine 191 studies between 2019 and 2024, organizing them into five key categories: 
\begin{itemize}
    \item \textbf{Accident Risk Prediction:} This involves predicting the likelihood of a traffic accident occurring.
    \item \textbf{Accident Frequency Prediction:} This task aims to estimate the number of traffic accidents within a specific time period.
    \item \textbf{Accident Severity Prediction:} This involves estimating how severe a traffic accident might be, considering factors like injuries, fatalities, and costs.
    \item \textbf{Accident Duration Prediction:} This task predicts the duration of traffic delays caused by accidents.
    \item \textbf{Statistical Modeling and Analysis:} This category encompasses studies that use statistical modeling and analytical techniques to analyze accident data, derive insights, or define predictive tasks. Examples include investigating the relationship between environmental factors and the occurrence of traffic accidents, as well as predicting accident hot spots.
\end{itemize}
We outline the strengths and weaknesses of these studies and carefully analyze how these methods manage the complexities and variations in traffic data and accident situations. We consolidate these findings to identify and discuss gaps in recent research allowing us to offer thoughtful recommendations for future work. 
Table \ref{tab:review_summary} showcases this study's contributions by summarizing its similarities and differences with previous work in this direction. 

\begin{table*}[ht]
    \centering
    \footnotesize
    \caption{A comparative summary between the recent surveys on traffic accident analysis and prediction and current study}
    \label{tab:review_summary}
    \begin{tabular}{@{}cccccc@{}}
        \toprule
        Study & \begin{tabular}{ccccc}
              & & Focus areas & \\
             \midrule
             Risk & Frequency & Severity & Duration & Analysis \\
             \end{tabular} & \begin{tabular}{@{}c@{}} Years \\ covered \end{tabular} & \begin{tabular}{@{}c@{}} Studies \\ considered \end{tabular} & \begin{tabular}{@{}c@{}} SOTA* \\ focus \end{tabular} & \begin{tabular}{@{}c@{}} Future work \\ discussed \end{tabular}\\
        \midrule
        \vspace{5pt}
        Silva et al. \cite{silva2020machine} & \begin{tabular}{ccccc} \hspace{11pt}\texttimes\hspace{17pt} & \hspace{7pt}\checkmark\hspace{17pt} & \hspace{15pt}\checkmark\hspace{13pt} & \hspace{15pt}\texttimes\hspace{13pt} & \hspace{15pt}\texttimes\hspace{13pt} \end{tabular} & 2003-2020 & 26 & Partially & \texttimes \\
        \vspace{5pt}
        Sarkar and Maiti \cite{sarkar2020machine} & \begin{tabular}{ccccc} \hspace{11pt}\checkmark\hspace{17pt} & \hspace{7pt}\texttimes\hspace{17pt} & \hspace{15pt}\checkmark\hspace{13pt} & \hspace{15pt}\texttimes\hspace{13pt} & \hspace{15pt}\texttimes\hspace{13pt} \end{tabular} & 1995-2019 & 232 & Partially & \checkmark \\
        \vspace{5pt}
        Angarita-Zapata et al.\cite{angarita2021bibliometric} & \begin{tabular}{ccccc} \hspace{11pt}\texttimes\hspace{17pt} & \hspace{7pt}\texttimes\hspace{17pt} & \hspace{15pt}\checkmark\hspace{13pt} & \hspace{15pt}\texttimes\hspace{13pt} & \hspace{15pt}\texttimes\hspace{13pt} \end{tabular} & 2011-2021 & 53 & Partially & Partially \\
        \vspace{5pt}
        Wen et al. \cite{wen2021applications} & \begin{tabular}{ccccc} \hspace{11pt}\texttimes\hspace{17pt} & \hspace{7pt}\texttimes\hspace{17pt} & \hspace{15pt}\checkmark\hspace{13pt} & \hspace{15pt}\texttimes\hspace{13pt} & \hspace{15pt}\texttimes\hspace{13pt} \end{tabular} & 2000-2020 & 88 & Partially & Partially \\
        \vspace{5pt}
        Santos et al. \cite{santos2022literature} & \begin{tabular}{ccccc} \hspace{11pt}\texttimes\hspace{17pt} & \hspace{7pt}\texttimes\hspace{17pt} & \hspace{15pt}\checkmark\hspace{13pt} & \hspace{15pt}\texttimes\hspace{13pt} & \hspace{15pt}\texttimes\hspace{13pt} \end{tabular} & 2001-2021 & 56 & Partially & \checkmark \\
        \vspace{5pt}
        Chand et al. \cite{chand2021road} & \begin{tabular}{ccccc} \hspace{11pt}\checkmark\hspace{17pt} & \hspace{7pt}\texttimes\hspace{17pt} & \hspace{15pt}\checkmark\hspace{13pt} & \hspace{15pt}\texttimes\hspace{13pt} & \hspace{15pt}\texttimes\hspace{13pt} \end{tabular} & 1998-2020 & 50+ & Partially & Partially \\
        \vspace{5pt}
        Ali et al. \cite{ali2024advances} & \begin{tabular}{ccccc} \hspace{11pt}\checkmark\hspace{17pt} & \hspace{7pt}\checkmark\hspace{17pt} & \hspace{15pt}\checkmark\hspace{13pt} & \hspace{15pt}\texttimes\hspace{13pt} & \hspace{15pt}\texttimes\hspace{13pt} \end{tabular} & 1997-2023 & 213 & Partially & \checkmark \\
        \vspace{5pt}
        \textbf{Current Study} & \begin{tabular}{ccccc} \hspace{11pt}\checkmark\hspace{17pt} & \hspace{7pt}\checkmark\hspace{17pt} & \hspace{15pt}\checkmark\hspace{13pt} & \hspace{15pt}\checkmark\hspace{13pt} & \hspace{15pt}\checkmark\hspace{13pt} \end{tabular} & 2019-2024 & 191 & Fully & \checkmark \\
        \bottomrule
    \end{tabular}
    \footnotesize{* SOTA: State of the art}
\end{table*}

The key contributions of this paper are summarized as follows:
\begin{itemize}
    \item \textbf{A comprehensive review of state-of-the-art studies}: This paper achieves two main objectives: 1) it conducts a thorough review of machine-learning-based studies for accident analysis and prediction; and 2) it focuses on the most recent advancements, specifically studies from the past five years. In pursuit of these goals, we have examined over 190 recent publications.
    \item \textbf{Covering various aspects of accident analysis and prediction}: We aim to explore multiple relevant dimensions, including predicting the risk, frequency, severity, and duration of accidents, as well as other analytical studies related to accident data.
    \item \textbf{Providing recommendations for future research}: This study offers guidance for future investigations, covering areas from data collection and model development to practical applications such as autonomous driving and real-time prediction systems.
\end{itemize}

The rest of the paper is organized as follows. Section~\ref{sec:preliminaries} covers the basics, as well as our research method. Section~\ref{sec:risk} reviews studies on predicting accident risk, followed by studies on accident frequency in Section~\ref{sec:frequency}. Section~\ref{sec:severity} discusses research on analyzing and predicting accident severity. Section~\ref{sec:duration} gives an overview of methods and challenges in predicting how long the impact of accidents last. Section~\ref{sec:analysis} explores statistical and analytical methods used to derive insights or define predictive tasks from accident data. Finally, Section~\ref{sec:future} outlines future research directions and Section~\ref{sec:conc} concludes the paper.

\section{Research Method and Preliminaries}
\label{sec:preliminaries}
We begin this section with describing core concepts. Next, we outline the research methodology, explaining the steps taken to review existing studies that use machine learning in accident analysis and prediction. Finally, we offer our insights into the papers reviewed. 

\subsection{Concepts}
Before we explore the related studies, this section clarifies a few key terms and concepts to understand their differences and similarities.

\begin{definition}[Traditional Models]
    Traditional machine learning models are algorithms that map an input feature vector $X$, representing measurable attributes such as weather conditions, traffic volume, road quality, and time of day, to a predictive output $y$. This output can indicate the likelihood of a traffic accident, its frequency within a certain area, or the severity of such events. These models are built upon established statistical and computational principles, employing techniques such as Random Forest (RF), Gradient Boosting Machines (GBM), Linear Regression, SVM, and Multi-layer Perceptron (MLP). They are foundational in the field and provide robust predictions by focusing on direct relationships within the data.
\end{definition}

\begin{definition}[Deep Neural Network Models]
    Deep Neural Networks (DNNs) represent an advanced class of machine learning models that incorporate multiple layers of neurons to process data through a deep structure of interconnected layers. Each layer refines and abstracts input features to discern complex patterns essential for understanding intricate scenarios like traffic dynamics. DNNs include various architectures tailored to specific data types and tasks, such as Convolutional Neural Networks (CNNs) for spatial data, Recurrent Neural Networks (RNNs) for temporal sequences, and Transformers for handling sequential data with enhanced focus and contextual awareness. These models excel in environments where relationships between data points are non-linear and highly variable, making them particularly effective in predicting nuanced outcomes in traffic accident scenarios.
\end{definition}

\begin{definition}[Statistical Modeling and Analysis]
    Statistical modeling, as used in this paper, refers to the application of both simple and advanced techniques to investigate correlations between environmental stimuli and aspects of traffic accidents such as occurrence, severity, or duration, and also to predict these events. This approach is distinguished from machine-learning-based methods by its use of techniques such as regression analysis, Poisson processes, and time-series modeling and prediction.
\end{definition}

\subsection{Research Methodology}
\begin{figure*}[ht!]
    \centering
    \includegraphics[width=0.9\linewidth]{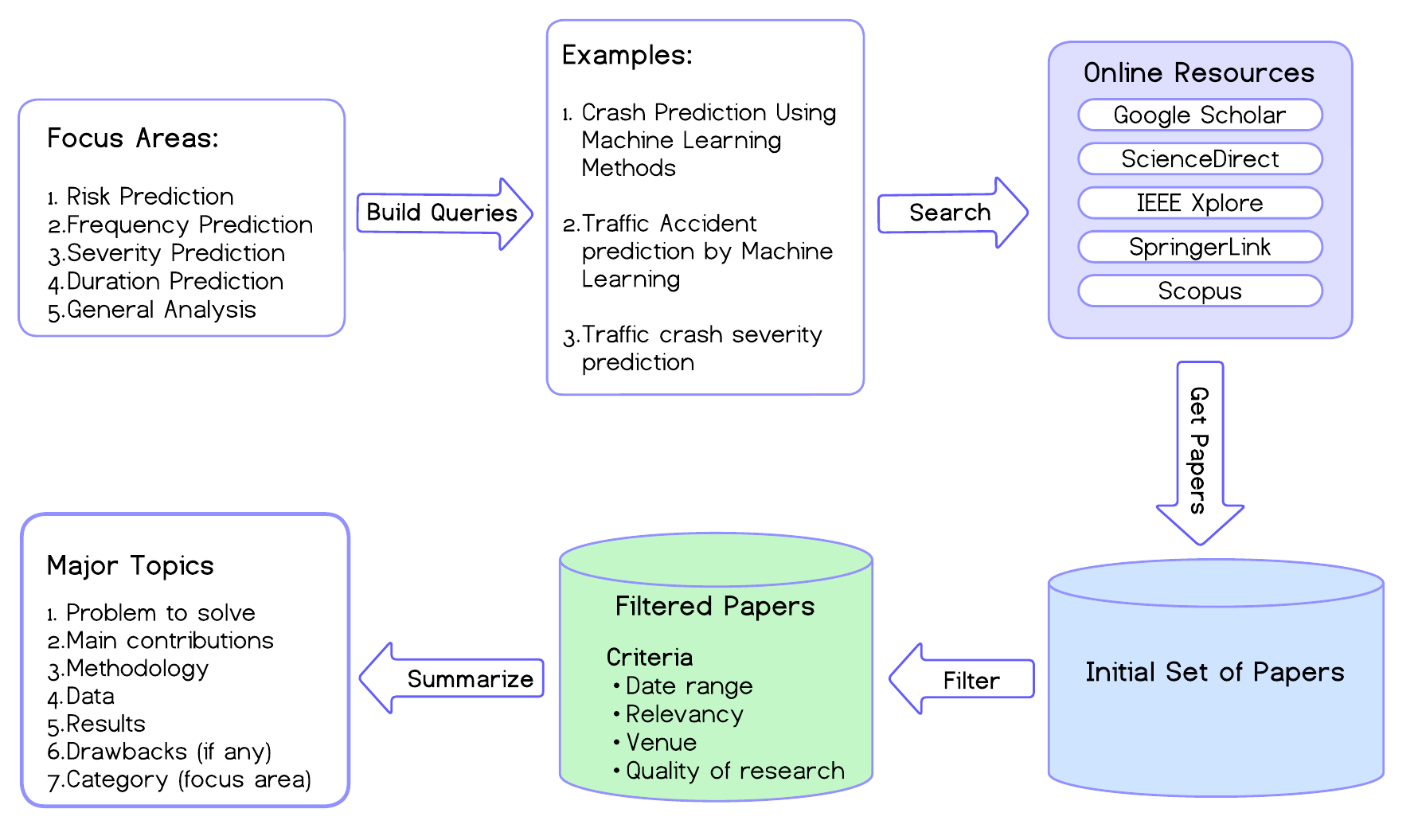}
    \caption{Our Process to Collect and Summarize Relevant Research Articles}
    \label{fig:process}
\end{figure*}

As Section~\ref{sec:introduction} describes, this paper focuses on five key areas: 1) predicting accident risk or occurrence, 2) predicting accident frequency, 3) predicting accident severity, 4) predicting post-accident impact duration, and 5) conducting general statistical modeling and analysis using accident data. Based on these categories, a straightforward process was developed to gather relevant research from the past five years, establishing a foundation to define the state-of-the-art in this domain. Figure~\ref{fig:process} illustrates this process, with the main steps outlined below.

\begin{itemize}
    \item \textbf{Focus areas}: This step involves defining the focus areas as outlined above.
    \item \textbf{Building search queries}: For each focus area, search queries are empirically developed to facilitate the discovery of related papers.
    \item \textbf{Searching through online resources}: Online resources such as Google Scholar\footnote{\hyperlink{Google Scholar}{https://scholar.google.com/}}, Science Direct\footnote{\hyperlink{Science Direct}{https://sciencedirect.com/}}, IEEE Xplore\footnote{\hyperlink{IEEE Xplore}{https://ieeexplore.ieee.org/}}, Springer Link\footnote{\hyperlink{Springer Link}{https://link.springer.com/}}, and Scopus\footnote{\hyperlink{Scopus}{https://www.scopus.com/}} were used to collect papers. Initial investigations indicated that these platforms should return most of the related research articles, supporting ``comprehensiveness'' of this study.
    \item \textbf{Collecting initial set of papers}: Over 250 research articles were collected between 2019 and 2024 using the online resources.
    \item \textbf{Filtering papers}: Each paper was reviewed to exclude those that were out of scope (due to relevancy or publication date), lacked publication venue information\footnote{Publications in well-known online resources such as \hyperlink{arXiv}{http://arxiv.org/} were not filtered out}, or where the quality of research was deemed low\footnote{Determined through the quality of the write-up and the methodology and results, upon careful review}. After filtering, 191 research articles remained.
    \item \textbf{Summarizing filtered papers}: The final step focuses on summarizing the remaining articles. The summary aims to extract the following information from each paper:
    \begin{itemize}
        \item Problem to solve: Identifies the main problem addressed by the paper.
        \item Main contributions: Highlights the primary contributions of the paper.
        \item Methodology: Describes the main methodology or model proposed by the paper.
        \item Data: Details the sources of data used in the paper's experiments.
        \item Results: Summarizes the major results and findings of the paper.
        \item Drawbacks: Discusses areas where the paper could improve or was lacking.
        \item Category: Specifies the focus area of the paper.
    \end{itemize}
\end{itemize}

\subsection{Overview of The Collected Papers}
\begin{figure}[h!]
    \centering
    \includegraphics[width=0.93\linewidth]{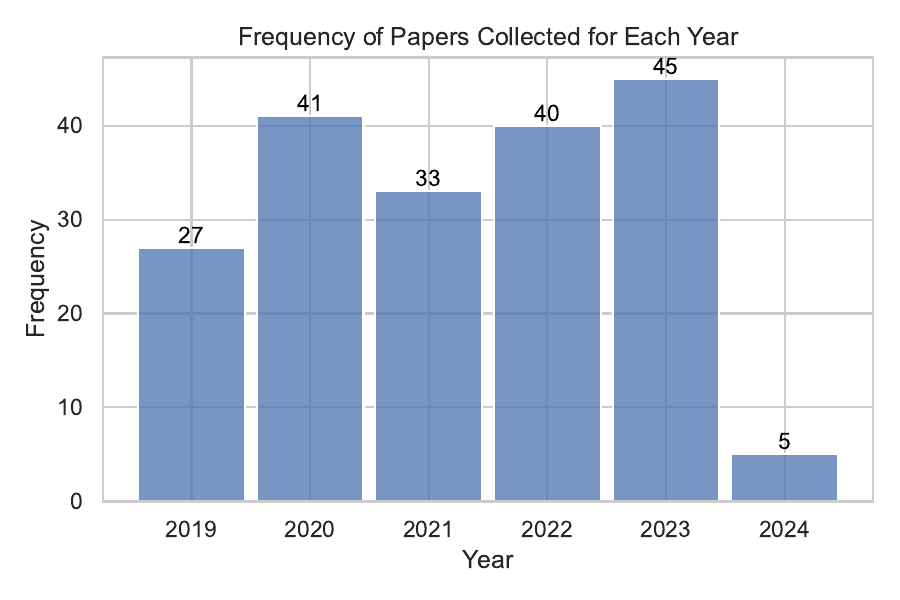}
    \caption{Yearly distribution of reviewed research papers}
    \label{fig:freq}
\end{figure}

This section provides a concise overview of the research papers collected. As previously mentioned, 191 research papers from the period 2019 to 2024 were assembled using various online tools. Figure~\ref{fig:freq} illustrates the yearly distribution of the papers. The collection aims to encompass literature over the past five years and includes several papers from 2024 that are pertinent to the research\footnote{It is worth noting that our collection encompasses all relevant papers available up until the end of March 2024.}.

\begin{figure}[h!]
    \centering
    \includegraphics[width=0.97\linewidth]{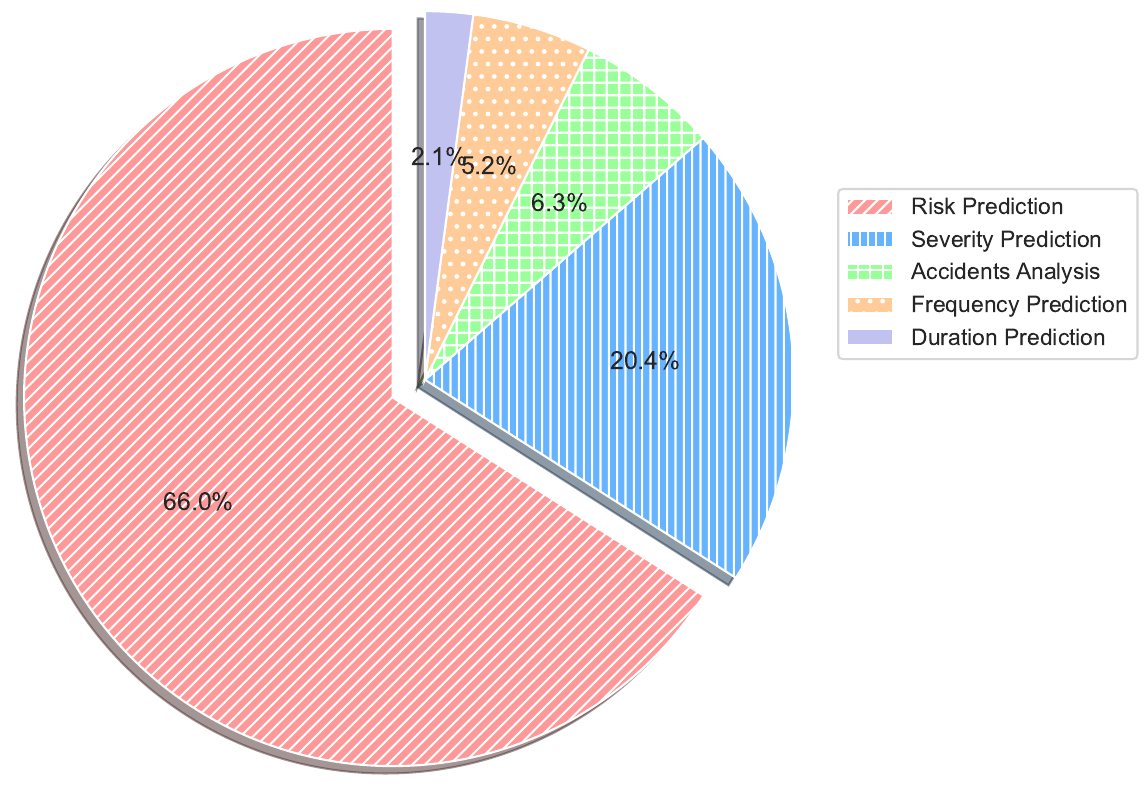}
    \caption{Distribution of different research categories that are reviewed in this study}
    \label{fig:categories}
\end{figure}

The distribution of papers across five categories is shown in Figure~\ref{fig:categories}. The data indicates that a significant number of papers focus on accident risk or occurrence prediction, with accident severity prediction ranking as the second most frequent category.

\begin{figure}[h!]
    \centering
    \includegraphics[width=0.9\linewidth]{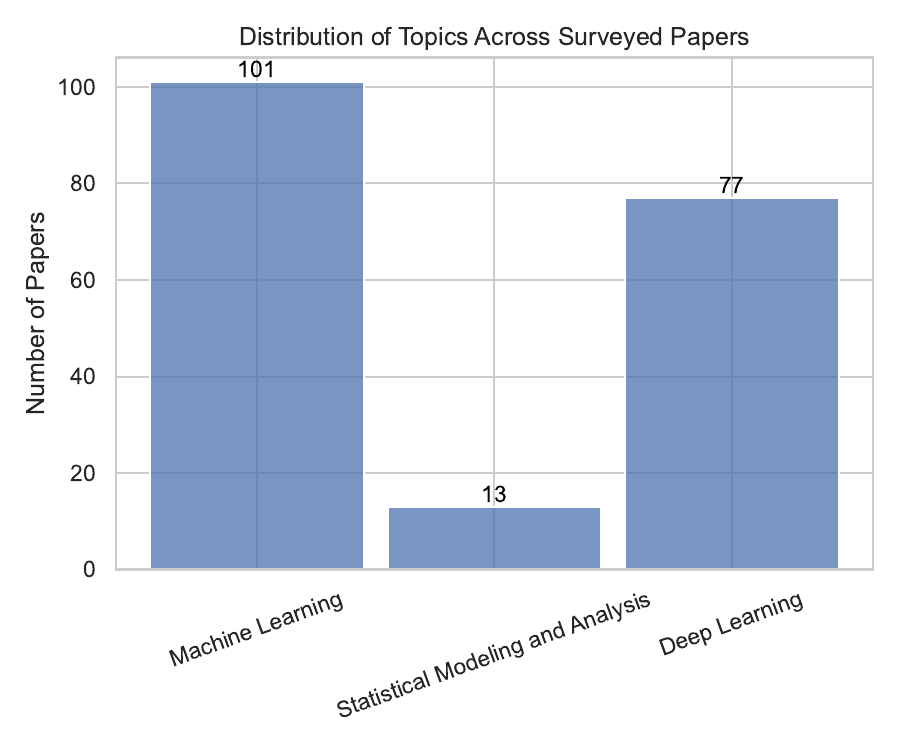}
    \caption{Distribution of major topics in the reviewed papers}
    \label{fig:topics}
\end{figure}

Next, Figure~\ref{fig:topics} summarizes the major topics in terms of the primary approaches employed in the reviewed papers. As the figure indicates, the majority of the papers fall under the ``machine learning'' topic, with ``deep learning'' as a closely following trend, reflecting the current state-of-the-art in the field.

\begin{figure}[h!]
    \centering
    \includegraphics[width=0.97\linewidth]{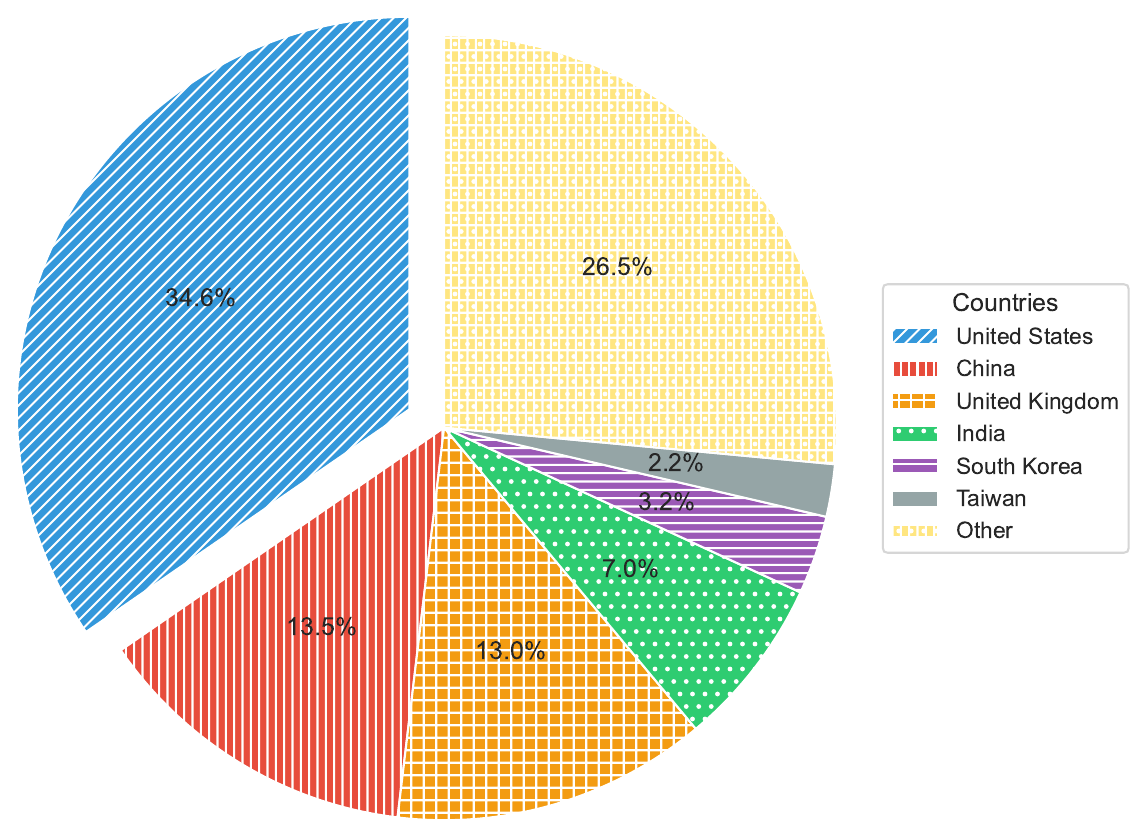}
    \caption{Geographical distribution of data sources from reviewed papers}
    \label{fig:countries}
\end{figure}

Lastly, Figure~\ref{fig:countries} presents the countries from which the input data originate, based on the reviewed papers. Notably, nearly a third of the studies utilize data from the United States, followed by China, the United Kingdom, and India. This distribution does not reflect the global prevalence of accidents (see WHO report 2023 \cite{WHO2023}), but rather the availability of public data used in these studies. This should encourage traffic authorities worldwide to enhance their processes, improving the accessibility of data for research aimed at boosting safety and reducing global accident rates.

\section{Accident Risk Prediction}
\label{sec:risk}

\subsection{Risk Prediction Using Traditional Machine Learning Models}
This section showcases the comprehensive body of research in the development and application of traditional machine learning models (Bayesian networks, ensemble learning, hybrid models, and spatiotemporal frameworks) to real-time traffic accident prediction. \cite{wu2019bayesian, chen2019novel, la2019development, wang2019expressway, dutta2019improving, zhao2019vehicle, dong2019roadway, ballamudi2019road, parsa2019real, wang2019crash2, wang2019crash, mor2020application, dabhade2020road, venkat2020machine, parra2020evaluating, lin2020real, fiorentini2020handling, abou2020proactive, zhai2020real, yassin2020road, hebert2020estimation, yan2020crash, mondal2020advancement, abou2020real, tang2020improving, peng2020examining, li2020traffic, wu2020novel, shi2021predicting, zhao2022prediction, alagarsamy2021prediction, bedane2021preventing, ariannezhad2021handling, guo2021study, ahsan2021machine, balfaqih2021accident, santos2021machine, sangare2021exploring, yang2022identification, erzurum2022prediction, atumo2022spatial, mehta2022road, augustine2022road, takale2022road, zhankaziev2022predicting, zheng2023traffic, gupta2022road, yeole2022prediction, zhang2022machine, dimitrijevic2022short, li2022hybrid, kaffash2022road, wei2022short, banerjee2022traffic, ahammad2023novel, yeole2023road, pourroostaei2023road, yang2023modeling, wei2023applying, ye2023data, zhao2023highway, fernando2023combining, krishna2023accident, almanie2023quantitative, infante2023prediction, li2023integrated, cui2023sparse}. 
The key takeaways from this body of work are that significant performance advances may be achieved through algorithmic innovations, but the primary limitation remains the data.

Zhao et al. \cite{zhao2019vehicle} introduce a real-time, accurate, and flexibly deployable accident risk prediction model utilizing Vehicular Ad-hoc NETwork (VANET) big data, employing a modified trichotomy AdaBoost algorithm. The model shows a 13.1\% higher accuracy and faster convergence compared to traditional methods, with an Area Under Curve (AUC) of 0.77. However, its one-hot encoding approach leads to lengthy feature vectors, increasing training time, and its effectiveness is only demonstrated on UK highway accident data, limiting its applicability to other environments. Hebert \cite{hebert2020estimation} develops high-resolution accident prediction models for Montreal by integrating multiple datasets. Using engineered features from these datasets, the models, particularly the balanced random forest, achieved a recall of 85\% with an AUC over 90\%. The model's heavy reliance on accident count, however, limits insights into other risk factors, and the 13\% false positive rate may hinder real-world applicability. Yassin \cite{yassin2020road} employs k-means clustering and random forest to predict road accident severity in Ethiopia, achieving 99.86\% accuracy. Driver experience and light conditions were among the top contributing factors. The study's reliance on a small, localized dataset and lack of comparison to advanced deep learning methods limit its generalizability and depth of analysis. Li et al. \cite{li2020traffic} propose a grey correlation analysis and optimized multivariable grey models for traffic accident prediction, achieving a low Mean Absolute Percentage Error (MAPE) of 1.48\% and surpassing other techniques in predicting Chinese traffic accidents. However, the model's consideration of only four variables and lack of evaluation in diverse geographical contexts may restrict its broader application. Santos et al. \cite{santos2021machine} analyze Setúbal, Portugal's road accident data to identify severity factors and predict hotspots using machine learning models. The models showed moderate accuracy and highlighted influential variables like motorcycle involvement. However, the study faced challenges with class imbalance and the specificity of findings to localized clusters, necessitating additional data for improved predictions. Sangare et al. \cite{sangare2021exploring} combine Gaussian mixture models and support vector classifiers to forecast traffic accident severity, achieving 88.5\% accuracy. This hybrid model leverages Gaussian Mixture Model (GMM) for feature extraction and SVM for discrimination, showing promise in severity classification. The study's limitations include the limited granularity of accident severity types and potential bias due to class imbalance during feature extraction. Banerjee et al. \cite{banerjee2022traffic} propose models to analyze and predict traffic accident risks, highlighting the utility of providing drivers with personalized risk scores and identifying accident hotspots. The lack of model validation on sample data and detailed operationalization strategies are significant shortcomings. Pourroostaei et al. \cite{pourroostaei2023road} utilize Apache Spark for robust accident prediction, achieving notable accuracy in predicting severity, casualties, and involved vehicles using traditional machine learning models. The reliance on an older UK dataset and the absence of certain environmental parameters are identified as limitations.

\begin{table*}[h!]
    \centering
    \footnotesize
    \caption{Summary of Other Notable Studies on Traffic Accident Risk Prediction Using Traditional Machine Learning Techniques}
    \begin{tabularx}{\linewidth}{|>{\hsize=.4\hsize}X|c|>{\hsize=.4\hsize}X|>{\hsize=1.8\hsize}X|>{\hsize=1.4\hsize}X|}
        \hline
        \rowcolor{lightblue} 
        \textbf{Study} & \textbf{Year} & \textbf{Main} \vfill \textbf{Approach} & \textbf{High-level Summary} & \textbf{Data Source} \\ \hline
            Dong et al. \cite{dong2019roadway} & 2019 & Machine Learning & Introduced a two-step traffic crash prediction method combining state-space model and support vector regression (SVR). & Tennessee DOT\footnote{Department of Transportation} data used for crash severity prediction on selected routes, 2010-2014. \\ \hline
            Ballamudi \cite{ballamudi2019road} & 2019 & Machine Learning & Explored machine learning models for predicting traffic accident injury severity, finding logistic regression most effective. & Analyzed road accidents using government data, social media, and onboard devices. \\ \hline
            Mor et al. \cite{mor2020application} & 2020 & Machine Learning & Develop an Accident Prediction Model for Haryana, India using linear regression, showing high accuracy in predictions. & Accident data from 1996-2016 in Haryana (India) from police, NCRB; includes casualties, accidents, GDP, road length, population, vehicles. \\ \hline
            Venkat et al. \cite{venkat2020machine} & 2020 & Machine Learning & Evaluated ensemble ML models for predicting road accident severity in NZ, highlighting Random Forest's effectiveness using SHAP analysis. & Datasets from Data.gov.uk, US-Accidents (2.25 million records) and UK police accidents (1.6 million records), 2000-2019. \\ \hline
            Parra et al. \cite{parra2020evaluating} & 2020 & Machine Learning & Explored explainable ML models using a large dataset to predict road traffic crashes, focusing on weather conditions. Utilized Random Forest and decision trees. & Data from the US-Accidents dataset. \\ \hline
            Dabhade et al. \cite{dabhade2020road} & 2020 & Machine Learning & Evaluated ensemble ML models for predicting road accident severity using NZ dataset, with RF and SHAP analysis. & Road accident data from Te Manatū Waka Ministry of Transport's CAS, New Zealand (2016-2020). \\ \hline
            Shi et al. \cite{shi2021predicting} & 2021 & Machine Learning & Developed a machine learning-based risk profiler for safer travel plans using geo-spatial data, achieving 31\% improved prediction accuracy. & Used static road, spatial-temporal features, accident records, and synthetic negative samples across six U.S. cities. \\ \hline
            Alagarsamy et al. \cite{alagarsamy2021prediction} & 2021 & Machine Learning & Developed a system using random forest and Gaussian distribution to identify and alert users about high-risk road areas. & Utilized Taihuyuan, South China and French National Interministerial Observatory traffic accident datasets. \\ \hline
            Ariannezhad et al. \cite{ariannezhad2021handling} & 2021 & Machine Learning & Improved real-time crash prediction using data mining on imbalanced data with random forest and AdaBoost models. & Data from ADOT traffic operation, ALISS database, and weather stations. \\ \hline
            Yeole et al. \cite{yeole2022prediction} & 2022 & Machine Learning & Developed an ANN-based model to predict traffic accidents, outperforming multiple linear regression, using real-world data. & Data from 887 accidents over six years in Pune, India, analyzed with weather, traffic, and road conditions. \\ \hline
            Gupta et al. \cite{gupta2022road} & 2022 & Machine Learning & Developed a deep learning model to predict traffic accidents using diverse factors, enhancing route planning and safety. & Used UK accident records, borough mappings, and economic index. \\ \hline
            Zheng et al. \cite{zheng2023traffic} & 2022 & Machine Learning & Developed non-parametric ML models (MLP, SVR, RF) to predict traffic conflicts, outperforming traditional safety methods. & Data from Surrey (Canada), Los Angeles, and Georgia (USA) on traffic crashes, volume, shockwave area, platoon ratio, shockwave speed. \\ \hline
            Augustine and Shukla \cite{augustine2022road} & 2022 & Machine Learning & Developed an Accident Prediction system using Machine Learning, with Random Forest showing the highest accuracy. & Dataset from Indian district government records (2018-2020) enhanced with synthetic non-accident data. \\ \hline
            Erzurum Cicek and Kamisli Ozturk \cite{erzurum2022prediction} & 2022 & Machine Learning & Explores one-class SVM's effectiveness in traffic accident prediction, outperforming traditional binary classifiers. & Fatal traffic accidents in Eskisehir, Turkey (2005-2012). \\ \hline
            Zhang et al. \cite{zhang2022machine} & 2022 & Machine Learning & Developed real-time traffic crash prediction models using RF, SVM, and XGBoost on statewide live traffic data. & Data sourced from HERE traffic database, CARE crash database, and HPMS highway infrastructure database. \\ \hline
            Cui et al. \cite{cui2023sparse} & 2023 & Machine Learning & Proposed a Sparse Spatio-Temporal Dynamic Hypergraph Learning framework to improve traffic accident prediction accuracy and interpretability. & Data from New York City and London on traffic accidents. \\ \hline
            Infante and Jacinto \cite{infante2023prediction} & 2023 & Machine Learning & Developed a predictive model using ML logistic regression to accurately forecast road traffic accidents. & Study utilized historical RTA data from EN10 highway in Portugal, collected from various national sources, focused on 2016-2022. \\ \hline
            Zhao et al. \cite{zhao2023highway} & 2023 & Machine Learning & Proposed a dynamic Bayesian network model for highway crash risk prediction with an 84.9\% accuracy, utilizing a random forest model. & Data from Digital Tachographs in South Korea for driving behavior, local government road geometry data, and urban mobility data in 2019. \\ \hline
            Krishna et al. \cite{krishna2023accident} & 2023 & Machine Learning & Developed a model to predict accident severity using machine learning, with random forest achieving 88.89\% accuracy. & Data on Tamil Nadu road accidents (2017-2020) from National Crime Records Bureau of India. \\ \hline
            Almanie \cite{almanie2023quantitative} & 2023 & Machine Learning & This paper introduces decision tree and naive Bayes models for predicting traffic accident severity, finding the decision tree superior. & Data from the US-Accidents dataset on Kaggle, focusing on Virginia accidents. \\ \hline
    \end{tabularx}
    \label{tab:risk_ml}
\end{table*}

Wu et al. \cite{wu2019bayesian} enhance real-time crash prediction by incorporating high-resolution freeway data from Los Angeles, selecting key variables via Random Forest, and employing a Bayesian network model, achieving improved sensitivity and accuracy over previous models. Limitations include potential non-transferability and unexplored variable discretization methods. Wang et al. \cite{wang2019crash2} propose a novel crash prediction method for urban expressways using floating car data and traffic platoon characteristics, with SVM models outperforming logistic regression. The study is limited by potential data biases and has only been validated on one expressway. Parsa et al. \cite{parsa2019real} compare SVM and Probabilistic Neural Network (PNN) models for real-time freeway accident detection using loop detector data, with PNN showing superior performance. The study did not explore deep learning models or a variety of data sources, limiting its generalizability. Dutta and Fontaine \cite{dutta2019improving} assess the relationship between crash occurrence and traffic flow quality, finding that speed data improves model predictions. Limitations include not considering temporal trends and the combination of various injury severity levels into one category. La Torre et al. \cite{la2019development} develop jurisdiction-specific accident prediction models for Italian freeways, showing alignment with Highway Safety Manual (HSM) coefficients but with under-prediction for high crash sites. The study's limitations include data assumptions and a focus on fatal and injury crashes only. Chen and Qin \cite{chen2019novel} introduce a real-time crash prediction method using simulated traffic data from a cell transmission model, showing a decrease in crash risk with variable speed limits. The study's reliance on accurate simulations and a small crash sample are notable limitations. Wu et al. \cite{wu2020novel} combine traditional and machine learning models for crash prediction, improving accuracy and stability. Limitations include a narrow focus on terrain types and a lack of detail on the parameter tuning process. Mondal et al. \cite{mondal2020advancement} compare RF and Bayesian Additive Regression Trees (BART) for weather-related crash severity prediction, with RF showing superior performance. The study lacks detailed model implementation information and generalizability. Yan et al. \cite{yan2020crash} address unobserved heterogeneity in crash data using dynamic road segmentation, finding improved prediction accuracy with Random Effects Negative Binomial (RENB) models. The study is limited by the accuracy of crash location data and a lack of detailed crash factors. Tang et al. \cite{tang2020improving} explore the use of transfer learning for crash prediction between jurisdictions, with Decision Tree algorithm based (DT-based) models showing better accuracy. The study's limitations include a lack of depth in model parameter analysis and limited geographic diversity. Zhai et al. \cite{zhai2020real} developed a real-time crash risk prediction model for foggy conditions, identifying key variables affecting crash risk. The study does not validate the models on separate test data and lacks a discussion on implementation challenges. Peng et al. \cite{peng2020examining} improve crash prediction accuracy by addressing the imbalanced data classification problem, with the RUSBoost\footnote{A machine learning algorithm that combines the concepts of Random Under-Sampling (RUS) and the Boosting technique to handle class imbalance effectively} model showing promising results. The study suggests further exploration of input variables and deep learning methods for improvement. Atumo et al. \cite{atumo2022spatial} utilize spatial statistics and Random Forest for crash hot spot prediction in Michigan, achieving high accuracy and highlighting the spatial dependence of crashes. Wei et al. \cite{wei2022short} analyze the combined effects of various factors on rural two-lane roadway crashes using eXtreme Gradient Boosting (XGBoost) models, identifying key influencing factors. The study is limited to rural two-lane roads and lacks certain roadway data. Dimitrijevic et al. \cite{dimitrijevic2022short} propose a framework for short-term crash risk prediction, with RF showing the best performance. However, the models did not achieve sufficient predictive performance for practical application. 

Wang et al. \cite{wang2019crash} address the limitations of traditional crash prediction models by proposing a method based on bivariate extreme value theory (EVT), integrating both perception-reaction and evasive action failures. Utilizing Unmanned Aerial Vehicle (UAV) captured video data and employing the Kanade-Lucas-Tomasi (KLT) technique and a multi-color fusion strategy, they achieve high vehicle detection rates and predict crash probabilities with notable accuracy. However, the study's limited sample size and the lack of comparison with advanced computer vision techniques are significant shortcomings. Fiorentini and Losa \cite{fiorentini2020handling} tackle the issue of imbalanced crash severity datasets in machine learning predictions. By employing random undersampling of the majority class, they demonstrate significant improvements in predicting minority severe and fatal crash classes across four ML algorithms. Their detailed analysis underscores the importance of class-specific metrics over overall accuracy. Lin and Li \cite{lin2020real} propose a hierarchical prediction method for traffic accident impact using crowdsourced data, distinguishing between different congestion levels and accident types. Their method, which includes both qualitative and quantitative predictions updated in real-time, shows promising results, particularly with neural networks. However, the limited variety of considered factors and the potential non-portability of city-specific models highlight areas for future improvement. Yeole et al. \cite{yeole2023road} develop a high-accuracy Artificial Neural Network (ANN) model for predicting road traffic accidents in mixed conditions, based on extensive parameters including weather, road, vehicle conditions, and driver behavior. 

A summary of the other notable studies on using traditional machine learning models to predict risk of traffic accidents is given in Table~\ref{tab:risk_ml}. 

\subsection{Risk Prediction Using Deep Neural Network Models}
This extensive body of research demonstrates the utilization of diverse deep learning models and methodologies, ranging from CNN and Long Short-term Memory (LSTM) to graph neural networks and attention mechanisms, for the purpose of enhancing the prediction accuracy and feature extraction in traffic accident analysis and forecasting \cite{zhao2019research,yuan2019real,theofilatos2019comparing,huang2019deep,chang2019deepcrash,moosavi2019accident,bao2019spatiotemporal,zhang2020traffic,naseer2020towards,bao2020uncertainty,zhou2020riskoracle,cai2020real,formosa2020predicting,huang2020highway,lin2020automated,li2020real,jiang2020long,singh2020deep,dinesh2021novel,basso2021deep,wang2021gsnet,lin2021intelligent,wang2021traffic,ali2021traffic,zhou2021efficient,yu2021deep,wang2022crash,hu2022data,jin2022deep,huang2022deep,man2022wasserstein,man2022transfer,huang2022traffic,gutierrez2022deep,park2022urban,zhao2022deep,li2022real,wakatsuki2022improving,zhang2022real,fang2022traffic,wu2023multi,jin2023prediction,zheng2023deep,li2023spatial,li2023research,karimi2023crashformer,trirat2023mg,azhar2023detection,zhang2023attention,kashifi2023deep,anik2023intformer,wang2023road,saravanarajan2023car,palit2023application,moosavi2023context,bhardwaj2023adaptive,liang2023unveiling,islam2023traffic, cho2023reinforcement}. 
The main takeaways are that some studies utilized existing deep neural networks like CNN, RNN, and Transformer models, while others used context-driven models to better handle heterogeneous input data. However, evaluating and testing these approaches on large, publicly accessible datasets is a noted challenge.

Bao et al. \cite{bao2019spatiotemporal} develop a citywide short-term crash risk prediction model using multi-source data and deep learning, focusing on spatiotemporal dependencies. They proposed a Spatiotemporal Convolutional LSTM Network (STCL-Net) to incorporate taxi Global Positioning System (GPS), weather, and road network data. Despite outperforming benchmarks, limitations include taxi GPS data biases and decreased performance at fine-grained predictions. Chang et al. \cite{chang2019deepcrash} propose DeepCrash, an Internet-of-Vehicles system for automatic high-speed crash detection and emergency notification using a modified Inception V3\footnote{The third version of the Inception CNN architecture.} model. Achieving 96\% accuracy and a 7-second notification time, it demonstrates the feasibility of such systems. However, the simulated crash testing and limited crash image dataset are significant shortcomings. 
Yuan et al. \cite{yuan2019real} aim to predict real-time crash risk at signalized intersections using an LSTM\footnote{Long Short Term Memory} model and the Synthetic Minority Over-sampling Technique (SMOTE) technique for balanced dataset generation. The model outperformed a conditional logistic model, showcasing the feasibility of real-time crash risk prediction with LSTM-RNN. 
Li et al. \cite{li2020real} develop a real-time crash risk prediction model for urban arterials using LSTM-CNN, integrating various data sources and SMOTE for data balancing. The model achieved high AUC and sensitivity, demonstrating the effectiveness of LSTM-CNN for crash prediction on complex urban arterials, despite the potential for further improvements and broader data inclusion. 
Formosa et al. \cite{formosa2020predicting} design a real-time traffic crash detection method using Region-based Convolutional Neural Network (R-CNN) and DNN models, achieving 94\% accuracy. This method integrates diverse data sources for Advanced Driver-Assistance Systems (ADAS) and autonomous vehicles, but its performance is limited to specific conditions and lacks real-time processing capabilities for large video data. 
Jiang et al. \cite{jiang2020long} introduce the LSTMDTR\footnote{LSTM based framework considering traffic data of different temporal resolutions for crash detection} model for crash detection using multi-resolution traffic data to capture dynamic pre-crash conditions, achieving higher accuracy than single-resolution models. The approach confirms the importance of considering traffic variations at different time scales. 
Basso et al. \cite{basso2021deep} utilize vehicle-by-vehicle data in deep learning models for crash prediction, achieving an AUC of 0.73 with the Deep Convolutional Generative Adversarial Network (DCGAN) technique. The study's spatial granularity and validation on a single highway segment are noted as its limitations. 
Li and Abdel-Aty \cite{li2022real} develop a TA-LSTM-CNN\footnote{Temporal Attention Long Short-Term Memory Convolutional Neural Network} model for real-time crash likelihood prediction using trajectory data fusion, outperforming benchmark models. However, the model's interpretability and transferability require further exploration. Palit and Osman \cite{palit2023application} predict near-crash events at an intersection using LiDAR\footnote{Light Detection and Ranging} data and a CNN+GRU\footnote{Gated Recurrent Unit} model, achieving high accuracy. The study's limited scope to one intersection and the absence of vehicle-pedestrian conflicts are considered weaknesses of the proposal. Anik et al. \cite{anik2023intformer} develop the IntFormer model for real-time crash likelihood prediction at intersections using connected vehicle data, showing promising results but high false alarm rates due to low CV\footnote{Connected Vehicles} market penetration. 
Kashifi et al. \cite{kashifi2023deep} present a short-term spatiotemporal crash prediction model using big traffic data in Paris, with DSHN\footnote{Deep Spatiotemporal Hybrid Network} outperforming other models. The study's city-specific focus and lack of exploration into different data window lengths are limitations. Li et al. \cite{li2023spatial} propose a method for predicting secondary crash locations using SSAE\footnote{Stacked Sparse Autoencoder}-LSTM, achieving promising spatial and temporal prediction accuracy. However, the limited secondary crash samples and potential data quality issues are key shortcomings. Karimi Monsefi et al. \cite{karimi2023crashformer} introduce CrashFormer, combining various transformer-based neural network architectures and achieving high performance in traffic accident risk prediction. 

Huang et al. \cite{huang2019deep} propose the DFN\footnote{Dynamic Fusion Network} framework, utilizing a context-aware embedding and hierarchical fusion network to address the challenges of heterogeneous data in traffic accident forecasting. Their approach showed significant improvement over competitive methods using New York City data, though the model's interpretability needs further elucidation. Moosavi et al. \cite{moosavi2019accident} introduce a real-time traffic accident prediction framework, leveraging a large-scale dataset and an LSTM-CNN architecture to utilize heterogeneous inputs and improve prediction over traditional and neural baselines. They evaluate their proposal on multiple cities in the United States, demonstrating superior results. Ali et al. \cite{ali2021traffic} create a real-time traffic monitoring framework using social media data, employing ontology and deep learning for event detection and analysis. Despite high accuracy, the manual effort in ontology construction and complexity with expanding concepts pose challenges. Zhao et al. \cite{zhao2022deep} enhance real-time traffic accident risk prediction with a CNN-RF\footnote{Random Forest with CNN Features} algorithm, showing high performance on a UK dataset. The need for larger and more diverse datasets, along with real-time performance testing, remains a shortcoming. Park and Hong \cite{park2022urban} enhance accident prediction accuracy by employing an MLP with dynamic traffic data features from Seoul. This approach integrates static road attributes like length and speed limits with dynamic variables such as traffic volume and the sun’s position, improving the prediction model's responsiveness to varying conditions. Gutierrez-Osorio et al. \cite{gutierrez2022deep} predict Bogota traffic accidents using social media and climate data, facing challenges in regions with sparse accident data. Incorporating accident type and severity could enhance the model's richness. Fang et al. \cite{fang2022traffic} introduce SSC-TAD \footnote{An acronym for Self-Supervised Consistency learning framework for Traffic Accident Detection.} for detecting traffic accidents from dashcam videos, achieving high performance. However, failure cases in complex scenarios and a lack of cause-effect modeling limit the approach. Cho et al. \cite{cho2023reinforcement} enhance the early and accurate prediction of traffic accidents from dashcam videos using deep reinforcement learning and a novel saliency\footnote{A saliency module in computer vision and machine learning identifies the most significant parts of an image by simulating aspects of human visual attention} module. Their DARC\footnote{Double Actors and Regularized Critics} model, leveraging double actor networks, outperforms previous methods in terms of prediction timeliness and accuracy. Nonetheless, the reliance on the saliency module's imperfect risk capture and limited dataset testing are noted limitations. 

\begin{table*}[h!]
    \centering
    \footnotesize
    \caption{Summary of Other Notable Studies on Traffic Accident Risk Prediction Using Deep Learning Techniques}
    \begin{tabularx}{\linewidth}{|>{\hsize=.4\hsize}X|c|>{\hsize=.4\hsize}X|>{\hsize=1.8\hsize}X|>{\hsize=1.4\hsize}X|}
        \hline
        \rowcolor{lightblue} 
        \textbf{Study} & \textbf{Year} & \textbf{Main} \vfill \textbf{Approach} & \textbf{High-level Summary} & \textbf{Data Source} \\ \hline
        Zhang et al. \cite{zhang2020traffic} & 2020 & Deep Learning & Improved traffic accident prediction accuracy using an LSTM-GBRT model, aiding in traffic management decisions. & Data from National Bureau of Statistics of China on traffic accidents, GDP, road and population metrics, 1997-2016. \\ \hline
        Naseer et al. \cite{naseer2020towards} & 2020 & Deep Learning & Framework for analyzing traffic accident data using big data and deep learning for hotspot prediction. & Data from UK Road Transport Authority on traffic accidents (2014-2016), involving extensive attributes and diverse collection methods. \\ \hline
        Dinesh \cite{dinesh2021novel} & 2021 & Deep Learning & Introduces RoadSafety app using AI, Deep Learning for real-time accident risk prediction, utilizing feed-forward neural network. & Used accident, weather, and landmark data; analyzed with neural networks, Pearson coefficient; integrated into "RoadSafety" iOS app. \\ \hline
        Zhou et al. \cite{zhou2021efficient} & 2021 & Deep Learning & Introduced a Fusion-Residual Predictive Network (FRPN) for early hazard warning in driving, enhancing defensive driving strategies. Utilized deep learning. & Data sourced from YouTube, driving recorders; divided into accident, normal videos for ``self-accident dataset.'' \\ \hline
        Hu et al. \cite{hu2022data} & 2022 & Deep Learning & Improved traffic crash prediction using ConvLSTM, capturing spatial-temporal characteristics more effectively than FC-LSTM. & Used historical crash data from Ningbo, China. \\ \hline
        Jin et al. \cite{jin2022deep} & 2022 & Deep Learning & Developed a hierarchical deep learning model addressing data imbalance and environmental factors for urban traffic risk estimation. & Data from Korea Transportation Safety Authority and local government DTG, for 2019 Daejeon city traffic. \\ \hline
        Huang et al. \cite{huang2022deep} & 2022 & Deep Learning & Introduced a Gated Graph Convolutional Multi Task model for predicting traffic accident risks using urban road images. & Data from NYC OpenData on traffic accidents, static road network characteristics, and urban road network images. \\ \hline
        Huang and Hooi \cite{huang2022traffic} & 2022 & Deep Learning & Introduced TRAVEL model improving traffic accident prediction by incorporating road network structure, outperforming GNN baselines. & US-Accidents dataset and OpenStreetMap road network data. \\ \hline
        Wakatsuki et al. \cite{wakatsuki2022improving} & 2022 & Deep Learning & Enhanced CNN model for predicting road accidents using traffic, time, and weather data. & Data from 44 vehicle detectors on Tomei Expressway, Japan, Gotemba to Tokyo, 2009-2018, with additional weather and calendar info. \\ \hline
        Wu et al. \cite{wu2023multi} & 2023 & Deep Learning & Introduced MADGCN, enhancing traffic accident prediction using dynamic graph convolutions and cost-sensitive learning. & Study used real-world data from NYC and PEMS-Bay, covering accidents, traffic flow, POIs, weather, and road attributes. \\ \hline
        Jin and Noh \cite{jin2023prediction} & 2023 & Deep Learning & Introduces a novel traffic accident prediction system using CNN and DNN, addressing data imbalance and estimating risk severity. & DTG data from OBU-equipped vehicles and local government road geometry data, in South Korea. \\ \hline
        Li and Luo \cite{li2023research} & 2023 & Deep Learning & Enhanced STGCN model for traffic accident risk prediction incorporating spatial, visual effects, and city landscape similarity. & Study utilized datasets on traffic accidents, taxi orders, POIs, road attributes, weather, and street view images for analysis. \\ \hline
        Liang et al. \cite{liang2023unveiling} & 2023 & Deep Learning & Proposes a self-supervised learning model using multiscale satellite imagery for fatal crash risk estimation. & Used satellite imagery from Texas metropolitan areas. \\ \hline
    \end{tabularx}
    \label{tab:risk_dl}
\end{table*}

Lin et al. \cite{lin2020automated} explore enhancing traffic incident detection using generative adversarial networks (GANs) to overcome the scarcity and imbalance of real-time incident data. They extracted spatial-temporal traffic flow features and generated synthetic incident samples to balance datasets, which improved detection rates and reduced false alarms. However, they only evaluated performance on highway data using an SVM model, and differences in occupancy features between real and synthetic samples were noted as limitations. Cai et al. \cite{cai2020real} tackle the challenge of real-time crash prediction with deep convolutional generative adversarial networks (DCGANs), aiming to balance datasets by generating synthetic crash data. Their approach outperformed traditional oversampling methods and improved sensitivity, specificity, and AUC metrics in machine learning models. The study was limited by its use of data from only one expressway over a single year and did not consider external factors or computational costs. Man et al. \cite{man2022transfer} focus on the transferability of real-time crash prediction models across different time periods and locations using GANs and transfer learning to address class imbalance. They demonstrated that transfer learning significantly enhances model transferability compared to direct transfers. Limitations include a limited number of crash cases and undersampling of non-crash cases due to computational constraints. Man et al. \cite{man2022wasserstein} introduce a Wasserstein Generative Adversarial Network (WGAN) for balancing datasets in real-time crash prediction, testing its efficacy against traditional oversampling techniques like SMOTE and ADASYN. Utilizing an extremely imbalanced dataset from the M1 Motorway in UK, the WGAN not only improved model sensitivity and AUC but also demonstrated significant enhancements in handling rare crash events effectively. 

Zhou et al. \cite{zhou2020riskoracle} introduce RiskOracle, a framework for real-time, minute-level traffic accident forecasting. It addresses sparse and imbalanced data challenges through a multi-task differential time-varying graph network and hierarchical region selection. The framework significantly outperforms baselines, but further validation across diverse regions and deeper component interaction analysis could enhance its robustness and transparency. Bao et al. \cite{bao2020uncertainty} propose an uncertainty-aware model for early traffic accident anticipation from dashcam videos. By integrating spatio-temporal relational learning and Bayesian neural networks, the model offers improved performance and interpretability. Wang et al. \cite{wang2021gsnet} develop GSNet\footnote{Learning Spatial-Temporal Correlations from Geographical and Semantic Aspects for Traffic Accident Risk Forecasting}, leveraging geographical and semantic modules to forecast traffic accident risk with attention to rare event challenges. Despite its effectiveness demonstrated on the NYC dataset, the absence of a diverse testing environment and a lack of visual analysis of spatial-temporal patterns can be identified as its shortcomings. Wang et al. \cite{wang2021traffic} create a multi-scale traffic accident risk forecasting model, employing a novel combination of CNNs, Graph Convolutional Networks (GCNs), and LSTM with an attention mechanism. The model excels in performance, but its exploration beyond two geographical scales and application to other domains remain as future work. Zhang and Guo \cite{zhang2023attention} present a model that uses attention mechanisms and dilated residual layers for traffic accident prediction, outperforming several methods. However, its dependency on comprehensive data and limited generalizability pose challenges for wider application. Zheng et al. \cite{zheng2023deep} focus on city-wide traffic accident forecasting, balancing accuracy across different risk regions. The model's granularity and balance in loss function are innovative, yet finer-grained analysis and the impact on high-risk prediction accuracy warrant further investigation. Bhardwaj et al. \cite{bhardwaj2023adaptive} introduce an adaptive framework for traffic accident risk prediction, combining graph and grid-based features with an attention mechanism. While achieving notable performance improvements, the exclusion of human mobility data suggests potential for further enhancement. Trirat et al. \cite{trirat2023mg} propose MG-TAR\footnote{Multi-View Graph Convolutional Networks for Traffic Accident Risk Prediction}, incorporating dangerous driving data into a multi-view graph-based model for accident risk prediction.

A summary of other notable studies on using deep-neural-network-based techniques to predict risk of traffic accidents is given in Table~\ref{tab:risk_dl}. 

\section{Accident Frequency Prediction}
\label{sec:frequency}

\subsection{Frequency Prediction Using Traditional Machine Learning Models}
This group of studies has focused on utilizing traditional machine learning models to predict traffic accident frequencies, in order to demonstrating the potential of these models to enhance road safety. These studies highlight the challenges of data imbalance and the need for detailed feature selection to improve prediction accuracy. They emphasize the critical role of machine learning in developing effective accident prediction and analysis frameworks \cite{hebert2019high,feng2020towards,macedo2022traffic,gataric2023predicting,fiorentini2023overfitting,hasan2023freeway}. Hebert et al. \cite{hebert2019high} devise a model for predicting vehicle collisions in Montreal with fine spatial and temporal granularity, employing Spark to integrate three datasets and generate both positive and negative examples, sampling only 0.1\% of the latter. Their feature engineering encompass meteorological conditions, road attributes, and time factors. They evaluate RF, Balanced RF (BRF)—an adaptation in Spark to undersample the majority class—and XGBoost, with BRF achieving an 85\% recall and 13\% false positive rate. Performance was comparable between BRF and RF, with XGBoost slightly inferior. The model heavily relies on historical accident data and temperature, facing limitations in causal inference due to this reliance and persistent class imbalance. Feng et al. \cite{feng2020towards} tackle road accident analysis in Albania, extracting 829 accident reports through Python and manually geotagging them to generate heatmaps via Folium, pinpointing high-risk zones like the Tirana-Durres highway. Despite revealing critical areas, the study's manual data processing and keyword-based analysis for Albanian proved laborious and constrained by the lack of comprehensive governmental data, hindering external validation. 

Fiorentini et al. \cite{fiorentini2023overfitting} employ Bayesian regularization (BR) backpropagation in an ANN to predict accidents on Italian four-lane divided roads, addressing small sample sizes and overfitting by incorporating a regularization term. This BR-ANN, novel in road safety analysis, predicts fatal and injury crashes across 236 road segments totaling 78 km, using inputs like road geometry, speed, traffic, and area type. It outperformed a gradient descent-trained ANN (GD-ANN) with lower overfitting, achieving a 0.726 determination coefficient ($R^2$) and Gaussian residuals distribution, suggesting its utility for small sample road safety analyses. Hasan et al. \cite{hasan2023freeway} evaluate variable speed limit (VSL) and variable advisory speed (VAS) as effective active traffic management strategies to harmonize traffic speeds based on real-time conditions. They develop short-term safety performance functions (SPFs) using traffic data and VSL/VAS operational information, which surpasses traditional models by considering temporal traffic characteristics. The Poisson log-normal model excelled across different traffic aggregations, confirming traffic volume and speed variability as crash predictors. VSL/VAS implementation reduces crashes by up to 26.14\% in peak traffic models, underscoring the strategies' efficacy for freeway safety improvement.

\subsection{Frequency Prediction Using Deep Neural Network Models}
The collective research advances traffic safety analytics by introducing high-resolution risk maps and spatio-temporal prediction frameworks. Additionally, it features innovative deep learning models for driver behavior analysis and accident prediction, leveraging extensive datasets and advanced neural network architectures \cite{he2021inferring, de2021new, wu2021mid, nippani2023graph}. He et al. \cite{he2021inferring} underscore the significant economic and human toll of traffic accidents, highlighting the potential of accident risk maps in monitoring and mitigating risks. They critique previous attempts for producing either low-resolution maps that miss critical details or relying on kernel density estimation prone to high variance. Addressing the challenge of creating high-resolution (5m) accident risk maps amidst sparse accident data, they introduce a deep learning model that fuses satellite imagery, GPS trajectories, road maps, and historical accident data. Their model, leveraging a deep neural network, generates a 5m resolution accident probability distribution map by establishing locational similarities through visual, traffic, and geographic data. Tested over 7,488 km\(^2\) across four US metro areas, their method surpasses existing models in both resolution and accuracy, attributing its success to the combination of diverse data sources. However, they caution against the model's complexity and the need for meticulous hyperparameter tuning to avoid overfitting. In parallel, Nippani et al. \cite{nippani2023graph} recognize the societal costs of traffic accidents and the need for improved predictive models. They spotlight the absence of comprehensive public datasets for traffic accident records, which hinders the evaluation of graph neural networks in accident prediction. To bridge this gap, they compile an expansive dataset containing over 9 million traffic accident records from eight US states over two decades. Utilizing this dataset, their evaluation of graph neural networks, particularly GraphSAGE\footnote{GraphSAGE is a framework designed for inductive representation learning on large-scale graphs.}, exhibits promising results with less than 22\% mean absolute error and over 87\% Area Under the Receiver Operating Characteristics (AUROC). They further enhance their model through multitask and transfer learning techniques, leveraging cross-state correlations and integrating traffic volume predictions. Despite achieving improvements with these methods, they acknowledge the limitations of simplified road network assumptions and the challenges posed by sparse labeling and potential under-reporting of accidents.

\vspace{5pt}
The main takeaways from the studies are that both traditional machine learning and deep learning models are useful for predicting traffic accident frequencies, but require careful data handling and model complexity to be effective. Machine learning excels at pinpointing accident likelihood, while deep learning can create detailed risk maps using a wider range of data. Key challenges include model complexity, needing a lot of data, and avoiding unreliable predictions. 
A summary of studies in the category of traffic accident frequency prediction is provided in Table~\ref{tab:freq}. 

\begin{table*}[h!]
    \centering
    \footnotesize
    \caption{Summary of Studies on Traffic Accidents Frequency Prediction}
    \begin{tabularx}{\linewidth}{|>{\hsize=.4\hsize}X|c|>{\hsize=.4\hsize}X|>{\hsize=1.8\hsize}X|>{\hsize=1.4\hsize}X|}
        \hline
        \rowcolor{lightblue} 
        \textbf{Study} & \textbf{Year} & \textbf{Main} \vfill \textbf{Approach} & \textbf{High-level Summary} & \textbf{Data Source} \\ \hline
        Hebert et al. \cite{hebert2019high} & 2019 & Machine Learning & Developed high-resolution accident prediction models for Montreal using big data analytics and the Balanced Random Forest algorithm. & Data from Montreal city, the Canadian government, and historical weather records. \\ \hline
        Feng et al. \cite{feng2020towards} & 2020 & Machine Learning & Developed a big data analytics platform using ML and DL for UK traffic accident analysis and prediction. & Compiled road accident data from Albanian online news using web scraping. \\ \hline
        He et al. \cite{he2021inferring} & 2021 & Deep Learning & Introduced a high-resolution accident risk map using deep learning on satellite and GPS data. & Data from satellite imagery, GPS trajectories, road maps, and accident history in United States. \\ \hline
        De Medrano and Aznarte \cite{de2021new} & 2021 & Deep Learning & Proposed a spatio-temporal deep learning framework for predicting traffic accidents in Madrid using a latent model. & Used Madrid's traffic, accident, and weather data from city's open data portal and municipal meteorological network for 2018. \\ \hline
        Wu and Hsu \cite{wu2021mid} & 2021 & Deep Learning & Developed a CNN-GRU fusion deep learning model to predict at-fault crash driver frequency, outperforming other approaches. & Traffic Police Corps, Taoyuan Police Department database, 2010-2018 fatal/injury crashes and citations. \\ \hline
        Macedo et al. \cite{macedo2022traffic} & 2022 & Statistical Modeling & Developed a GIS-based accident prediction model for rural highways, utilizing the GEE model, focused on Brazil. & Accident records from Federal Highway Police, geometric highway data from DNIT and satellite imagery. \\ \hline
        Nippani et al. \cite{nippani2023graph} & 2023 & Deep Learning & Constructed a large-scale traffic accident dataset, significantly improving prediction accuracy with GraphSAGE and multitask learning. & 9 million accident data collected from State DOTs, OpenStreetMap; included road, weather, and traffic data. \\ \hline
        Gataric \cite{gataric2023predicting} & 2023 & Deep Learning & This study developed two ANN models to predict traffic accidents and their severity on common roads. & Data from Serbia and Bosnia and Herzegovina's road agencies, collected between 2017 and 2021. \\ \hline
        Fiorentini et al. \cite{fiorentini2023overfitting} & 2023 & Deep Learning & Implemented Bayesian regularization backpropagation in ANN for accurate road accident prediction on Italian roads, overcoming overfitting. & 236 road elements totaling 78 km, including details like alignment, geometry, speed, traffic, and area type. \\ \hline
        Hasan et al. \cite{hasan2023freeway} & 2023 & Statistical Modeling & Developed short-term safety performance functions using Poisson log-normal model, highlighting VSL/VAS effectiveness in reducing crash frequency. & High-resolution traffic detector data and VSL/VAS operational data. \\ \hline
    \end{tabularx}
    \label{tab:freq}
\end{table*}

\section{Accident Severity Prediction}
\label{sec:severity}

\subsection{Severity Prediction Using Traditional Machine Learning Models}
This collection of studies explores the application of various machine learning techniques, including random forest, decision trees, neural networks, and ensemble methods, to predict and analyze traffic accident severity, consistently highlighting the effectiveness of these models in enhancing prediction accuracy \cite{lee2019model, sarkar2019application, almamlook2019comparison, labib2019road, wahab2020severity, kashyap2020traffic, geyik2020severity, paul2020prediction, pradhan2020modeling, mansoor2020crash, jadhav2020road, assi2020predicting, assi2020traffic, ijaz2021comparative, yan2021single, malik2021road, evwiekpaefe2021predicting, iveta2021prediction, ghasedi2021prediction, niyogisubizo2021comparative, yang2022predicting, kumar2022road, koramati2023development, yan2022traffic, li2023predicting, yang2023prediction, obasi2023evaluating, cceven2024traffic, peng2024transportation}. 

Sarkar et al. \cite{sarkar2019application} explore ML techniques—SVM and ANN—optimized by Genetic Algorithms (GA) and Particle Swarm Optimization (PSO) to predict occupational incidents. They stress parameter tuning's importance for effective predictions and highlight rule extraction's significance to understand accident factors' inter-relationships. Their GA and PSO optimized models achieve 90.67\% prediction accuracy, surpassing traditional methods. Despite successful root cause identification through nine rules, the study is limited by a single organization's dataset and extensive manual data pre-processing. 
Almamlook et al. \cite{almamlook2019comparison} demonstrate that machine learning models, especially Random Forest (RF), can effectively predict traffic accident severity in Michigan, USA. RF outperformed Logistic Regression, Naive Bayes, and AdaBoost with an accuracy of 75.5\%. The study's limitation lies in its focus on Michigan data and the exclusion of factors such as driver behavior, suggesting the need for a broader dataset and improved modeling techniques. Assi et al. \cite{assi2020predicting} improve crash severity prediction using machine learning, with the SVM-Fuzzy C-Means clustering model showing the best performance (74.2\% accuracy). The study highlighted the potential of integrating clustering techniques but was critiqued for not addressing class imbalance and the challenge of applying the models in countries with less comprehensive datasets. Wahab and Jiang \cite{wahab2020severity} evaluate MLP, Partial Decision Tree (PART), and SimpleCart\footnote{simplified version of the CART (Classification and Regression Trees) algorithm} models for predicting motorcycle crash severity in Ghana, finding SimpleCart to be the most accurate (73.81\%). The study identified significant factors such as crash location and time, although it did not explore other machine learning approaches or deeper data analysis. Ghasedi et al. \cite{ghasedi2021prediction} focus on identifying factors influencing crash severity and develop predictive models using ANN and logistic regression, with ANN showing superior performance. The research was constrained by limited pedestrian data and the absence of a comparative analysis with other machine learning approaches. Obasi and Benson \cite{obasi2023evaluating} critique traditional statistical methods for accident severity prediction and highlight the challenges of machine learning applications. Using UK data, they found RF and Logistic Regression to be the most accurate (87\%). However, the study noted the lack of detailed casualty information and the limitations of single-model approaches.

Lee et al. \cite{lee2019model} demonstrate that machine learning models, particularly RF, ANN, and Decision Trees, outperform traditional statistical methods in predicting traffic accident severity by capturing complex nonlinear relationships without pre-assumptions. Their study utilize three datasets encompassing road geometry, rainfall data, and accident records, with RF showing the highest accuracy. However, the study faced limitations such as a small dataset not accounting for diverse road conditions and lacked independent validation and in-depth interpretation of RF predictions. Yan and Shen \cite{yan2022traffic} address the trade-off between prediction accuracy and model interpretability in traffic accident severity prediction by employing Bayesian optimized Random Forest (BO-RF). Their methodology include extensive data preprocessing and hyperparameter optimization, leading to superior performance in terms of recall, F1 score, and AUC. Despite its robustness, the model's limitations include potential bias due to imbalanced training data, a limited number of input features, and the subjective nature of model interpretation methods. Yang et al. \cite{yang2022predicting} explore the impact of road and environmental factors on freeway crash severity through an XGBoost-Bayesian network model, achieving a prediction accuracy of 89.05\%. This study highlighted the significance of considering feature interactions for freeway safety management. However, the research is limited by a small crash data sample size and a restricted set of road condition variables. Çeven et al. \cite{cceven2024traffic} focus on classifying traffic accident severity in urban areas using DT based ensemble methods. The study highlights driver fault as a significant factor affecting accident severity, with high F1 scores indicating model effectiveness. Peng et al. \cite{peng2024transportation} investigate the changing influences on traffic accident severity across different stages of the COVID-19 pandemic using a RF model. The study provides insights into the temporal variation of influential factors, such as traffic signals and weather conditions. Limitations include the geographic restriction to Illinois, USA, and the absence of driver and vehicle-specific variables in the analysis.

The main takeaways are that machine learning techniques are effective in predicting traffic accident severity, emphasizing the importance of feature interactions and data preprocessing. However, challenges include handling imbalanced data, extensive preprocessing requirements, and ensuring model interpretability and generalizability. Limitations often stem from dataset constraints, geographic focus, and the exclusion of certain influential variables. 
A summary of some of the other notable studies in this category are listed in Table~\ref{tab:sev_ml}

\begin{table*}[h!]
    \centering
    \footnotesize
    \caption{Summary of the Other Notable Studies on Traffic Severity Prediction using Machine Learning Techniques}
    \begin{tabularx}{\linewidth}{|>{\hsize=.4\hsize}X|c|>{\hsize=.4\hsize}X|>{\hsize=1.8\hsize}X|>{\hsize=1.4\hsize}X|}
        \hline
        \rowcolor{lightblue} 
        \textbf{Study} & \textbf{Year} & \textbf{Main} \vfill \textbf{Approach} & \textbf{High-level Summary} & \textbf{Data Source} \\ \hline
        Kashyap et al. \cite{kashyap2020traffic} & 2020 & Machine Learning & Developed a framework using ML models like Random Forest and Logistic Regression, achieving 87\% accuracy in predicting traffic accident severity. & Data from UK traffic accident records (2005-2014), accident and vehicle information databases. \\ \hline
        Geyik and Kara \cite{geyik2020severity} & 2020 & Machine Learning & Developed models to predict UK traffic accident severity using MLP, Decision Tree, Random Forest, and Naive Bayes. & Used UK's stats19 traffic accident data (2010-2012). \\ \hline
        Paul et al. \cite{paul2020prediction} & 2020 & Machine Learning & Developed a combined accident prediction and severity model using machine learning, achieving best results with Decision Tree. & Collected road accident datasets for Bangladesh from BRTA and PPRC, covering injuries, severity, and various casualties over years. \\ \hline
        Pradhan et al. \cite{pradhan2020modeling} & 2020 & Machine Learning & Explored various models like LR, ANNs, SVMs, and Bayesian methods for traffic accident severity analysis. & Data from police reports (2009-2015) on 1,138 accidents along Malaysia's North-South Expressway. \\ \hline
        Mansoor et al. \cite{mansoor2020crash} & 2020 & Machine Learning & Proposed a two-layer ensemble machine learning model to accurately predict road traffic crash severity. & Department of Transport, Great Britain, for road intersection crashes, 2011-2016. \\ \hline
        Jadhav et al. \cite{jadhav2020road} & 2020 & Machine Learning & This paper improves road accident severity classification in Bangladesh using machine learning models, best performed by AdaBoost. & Data from YouTube; surveillance videos at 30 fps, trimmed to 20 seconds; diverse ambient conditions. \\ \hline
        Malik et al. \cite{malik2021road} & 2021 & Machine Learning & Developed a framework using ML algorithms like Random Forest to predict and analyze crash severity. & Data from the UK Department of Transport, covering 122,636 road accidents. \\ \hline
        Evwiekpaefe and Umar \cite{evwiekpaefe2021predicting} & 2021 & Machine Learning & Utilized data mining algorithms to classify and identify the significant causes of Road Traffic Crashes in Nigeria. & Data from Federal Road Safety Corps, Nigeria, covering 1580 crash cases from 2016-2018, with 26 attributes analyzed. \\ \hline
        Iveta et al. \cite{iveta2021prediction} & 2021 & Machine Learning & Predicts road accident severity using machine learning with a multiclass classification model on normalized datasets. & Data from UK government's Road Traffic Accidents and Safety Statistics. \\ \hline
        Niyogisubizo et al. \cite{niyogisubizo2021comparative} & 2021 & Machine Learning & Utilized RF, MNB, KC, KNN for predicting road crash severity, highlighting RF's superior accuracy and feature importance analysis. & Used Victoria, Australia's road accident data from 2015-2020, provided by the Department of Transport. \\ \hline
        Kumar and Santosh \cite{kumar2022road} & 2022 & Machine Learning & Predictive model using logistic regression, decision tree, and random forest for forecasting traffic accident severity. & Data from the US-Accidents dataset. \\ \hline
        Koramati et al. \cite{koramati2023development} & 2022 & Deep Learning & Developed ANN models for urban crash prediction in India using Hyderabad police data, highlighting sensitive crash factors. & Road crash database from police records: 7464 total (2015-2019), 1714 total (2019) crashes analyzed. \\ \hline
        Li et al. \cite{li2023predicting} & 2023 & Machine Learning & Study enhances traffic accident severity prediction on mountain freeways using ML models like SVM, DTC, AdaBoost, and RF feature selection. & Real-time traffic, weather data, and accident data from 649 traffic accidents on a Yunnan freeway, China, from June 2019 to December 2020. \\ \hline
        Yang et al. \cite{yang2023prediction} & 2023 & Machine Learning & Innovates traffic accident severity prediction in China using Random Forest on augmented accident feature set. & Data from Chinese National Automobile Accident Investigation System (2018-2020). \\ \hline
    \end{tabularx}
    \label{tab:sev_ml}
\end{table*}

\subsection{Severity Prediction Using Deep Neural Network Models}
This compendium of studies showcases advancements in traffic accident severity prediction, employing a diverse array of deep learning techniques, ranging from novel model integrations and comprehensive frameworks to the utilization of extensive datasets and innovative approaches for enhanced prediction accuracy and road safety \cite{zheng2019traffic, manzoor2021rfcnn, vaiyapuri2021traffic, rahim2021deep, ma2021analytic, ponduru2023road, oliaee2023using, sattar2023transparent, khan2023novel, alhaek2024learning}. 
Zheng et al. \cite{zheng2019traffic} introduce a novel convolutional neural network model, TASP-CNN\footnote{traffic accident’s severity prediction-convolutional neural network}, for improving traffic accident severity prediction by transforming feature relationships into gray images. Their model outperformed several traditional statistical and machine learning models with a micro F1-score of 0.87. However, the model's performance could potentially be further enhanced by extending the hyperparameter tuning beyond 100 epochs and conducting more comprehensive ablation studies\footnote{An ablation study investigates the performance of an AI system by removing certain components or input features to understand the contribution of the component to the overall system.}. Manzoor et al. \cite{manzoor2021rfcnn} develop the RFCNN\footnote{An ensemble of machine learning and deep learning models by combining Random Forest and Convolutional Neural Network called RFCNN for the prediction of road accident severity} model that combines RF and CNN using soft voting, showing significant improvements in predictive accuracy when utilizing the 20 most influential features. While the model demonstrated high precision and recall, its complexity and potential lack of generalizability to other datasets were noted as limitations. Ma et al. \cite{ma2021analytic} present a comprehensive framework that employs multiple machine learning techniques, including CatBoost and stacked sparse autoencoders (SSAE), for predicting the severity of traffic accident injuries using UK data. This method notably enhances the accuracy by integrating both spatial and temporal data analysis, effectively predicting severe injuries in different regions. Despite its potential, the framework's limitations include rather limited analysis of vulnerable road user (VRU)-related data and varied performance across different spatial clusters. Oliaee et al. \cite{oliaee2023using} utilize Natural Language Processing (NLP) techniques with a fine-tuned Bidirectional Encoder Representations from Transformers (BERT) model to classify crash injury severity from narratives, achieving high accuracy. Alhaek et al. \cite{alhaek2024learning} propose a model combining CNN, Bidirectional LSTM (BiLSTM), and attention mechanisms for accident severity prediction, demonstrating effectiveness across different datasets. The study acknowledged the potential variability in performance across regions and the need for improved model interpretability.

While existing studies in this category often rely on complex deep neural network models, advancements in the machine learning field indicate potential for further improvement. Employing other innovative, sophisticated models and incorporating more diverse datasets could better demonstrate the effectiveness of these approaches. 
A summary of studies in the category of traffic accident severity prediction using deep learning techniques is provided in Table~\ref{tab:sev_dl}. 
\begin{table*}[h!]
    \centering
    \footnotesize
    \caption{Summary of Studies in Traffic Accident Severity Prediction Using Deep Learning Techniques}
    \begin{tabularx}{\linewidth}{|>{\hsize=.4\hsize}X|c|>{\hsize=.4\hsize}X|>{\hsize=1.8\hsize}X|>{\hsize=1.4\hsize}X|}
        \hline
        \rowcolor{lightblue} 
        \textbf{Study} & \textbf{Year} & \textbf{Main} \vfill \textbf{Approach} & \textbf{High-level Summary} & \textbf{Data Source} \\ \hline
            Zheng et al. \cite{zheng2019traffic} & 2019 & Deep Learning & Developed a novel TASP-CNN model improving traffic accident severity prediction by converting data into gray images. & Leeds City Council provided 21,436 traffic accident records (2009–2016). \\ \hline
            Manzoor et al. \cite{manzoor2021rfcnn} & 2021 & Deep Learning & Introduced RFCNN combining Random Forest and CNN to predict road accident severity, achieving high accuracy. & Used the USA-accidents dataset (Feb 2016 - Jun 2020) from Kaggle. \\ \hline
            Vaiyapuri and Gupta \cite{vaiyapuri2021traffic} & 2021 & Deep Learning & Developed a model for predicting traffic accident severity in India using ML and DL models, where MLP performed best. & Data from Road Crashes Repository and police records in Muzaffarnagar, India, including 2203 road accident records. \\ \hline
            Rahim and Hassan \cite{rahim2021deep} & 2021 & Deep Learning & Proposes a CNN deep learning model with custom f1-loss for improved traffic crash severity prediction. & Data on Louisiana traffic crashes (2014-2018). \\ \hline
            Ma et al. \cite{ma2021analytic} & 2021 & Deep Learning & Developed a comprehensive SSAE-based deep learning framework for predicting injury severity in traffic accidents, utilizing CatBoost and k-means clustering. & Data on traffic accidents and vehicles from the UK Department for Transport (2011-2016).  \\ \hline
            Ponduru \cite{ponduru2023road} & 2023 & Deep Learning & Developed deep learning models to predict accident severity using driver, road, vehicle, and environmental factors. & Dynamic Tachometer Graphs from vehicles in Korea, tracking operation data like speed and location. \\ \hline
            Oliaee et al. \cite{oliaee2023using} & 2023 & Deep Learning & Utilized BERT to classify traffic injury types from 750,000+ crash reports with 84.2\% accuracy and AUC of 0.93. & Data from 740,736 Louisiana crash reports (2011-2016).  \\ \hline
            Sattar et al. \cite{sattar2023transparent} & 2023 & Deep Learning & This paper models motor vehicle crash injury severity using MLP, MLP with embeddings, and TabNet, highlighting TabNet's feature importance for highway safety. & Data on GB crashes (2011-2016) included driver, vehicle, roadway, and crash attributes, aggregated severity levels. \\ \hline
            Khan and Ahmed \cite{khan2023novel} & 2023 & Deep Learning & Developed a ResNet18-based model using DeepInsight for predicting weather-related crashes on mountainous freeways with high accuracy. & The data source is unclear.  \\ \hline
            Alhaek et al. \cite{alhaek2024learning} & 2024 & Deep Learning & Proposes a CNN-BiLSTM model with attention mechanisms for predicting traffic accident severity, enhancing road safety. & UK traffic accident dataset, including location, date, time, vehicle types, and accident severity, used for data cleaning study. \\ \hline
    \end{tabularx}
    \label{tab:sev_dl}
\end{table*}

\section{Accident Duration Prediction}
\label{sec:duration}
This series of studies investigates the enhancement of traffic incident duration prediction, employing a range of machine learning and deep learning models including SVM, Gaussian Process Regression (GPR), Restricted Boltzmann Machines (RBMs), and advanced text and multi-modal data analysis techniques \cite{hamad2020predicting, li2020deep, chen2022traffic, chen2024traffic}. 
The main objective of these studies is to predict the duration of impact caused by traffic accidents using a range of contextual variables, such as weather conditions and traffic flow.
Hamad et al. \cite{hamad2020predicting} explore various machine learning models for predicting traffic accident duration. They experimented with regression trees, SVMs with different kernels, ensemble methods, Gaussian processes, and simple neural networks. The study found that complex regression trees outperformed other tree models, and the quadratic SVM showed high accuracy among SVMs. The GBM ensemble model was noted for its efficiency and accuracy. The Matern 5/2 kernel was the best among GPR models, and a simple ANN with 10 neurons was most accurate among neural network models. The models' interpretability and online performance were not assessed, and the study did not consider temporal changes over the years, suggesting the need for time-specific models.
Li et al. \cite{li2020deep} propose a deep fusion model to predict the duration of traffic accidents, considering both the characteristics of the accidents and the spatial-temporal correlations of traffic flow. The model uses stacked RBMs to handle different types of variables and a joint layer for feature fusion. Tested on I-80\footnote{The Interstate 80 (I-80) freeway dataset} data, the model outperformed benchmarks, showing that integrating various features improves prediction accuracy. This approach allows for a comprehensive analysis of traffic accident and flow data, highlighting the importance of feature fusion in predicting traffic accident durations.
Chen and Tao \cite{chen2022traffic} develop a model to predict traffic accident duration by integrating structured accident data with unstructured text from accident reports. They preprocessed text data, used Term Frequency-Inverse Document Frequency (TF-IDF) for feature vectors, and employed a RF algorithm for feature selection. Combining text features with structured data in ensemble models like XGBoost and RF improved prediction accuracy. Key textual factors included vehicle and accident types, highlighting the importance of text data in duration prediction. However, the study's limited sample size and focus on overall prediction rather than stages or short-range forecasting accuracy were noted as limitations.

A summary of studies in the category is provided in Table~\ref{tab:dur}. 

\begin{table*}[h!]
    \centering
    \footnotesize
    \caption{Summary of Studies on Accident Duration Prediction}
    \begin{tabularx}{\linewidth}{|>{\hsize=.4\hsize}X|c|>{\hsize=.4\hsize}X|>{\hsize=1.8\hsize}X|>{\hsize=1.4\hsize}X|}
        \hline
        \rowcolor{lightblue}
        \textbf{Study} & \textbf{Year} & \textbf{Main} \vfill \textbf{Approach} & \textbf{High-level Summary} & \textbf{Data Source} \\ \hline
        Hamad et al. \cite{hamad2020predicting} & 2020 & Machine Learning & Explores machine learning models, notably SVM and GPR, for accurate traffic incident duration prediction. & Data from the Transtar operators' database including 146,573 incidents with 74 attributes, modified for prediction analysis. \\ \hline
        Li et al. \cite{li2020deep} & 2020 & Deep Learning & Introduced a deep fusion model using RBMs to predict traffic accident durations considering spatial-temporal correlations. & Data from Highway Safety Information System (HSIS) and Caltrans Performance Measurement System (PeMs) collected on I-80 in US.  \\ \hline
        Chen and Tao \cite{chen2022traffic} & 2022 & Machine Learning & This paper innovates in traffic accident duration prediction by leveraging text data with an improved RF model. & Data from 22,497 traffic accident samples in Shaanxi, China (Jan 2020-Apr 2021), including structured and text descriptions. \\ \hline
        Chen et al. \cite{chen2024traffic} & 2024 & Deep Learning & Developed a Word2Vec-BiGRU-CNN model enhancing traffic accident duration prediction using multi-modal data analysis. & Traffic accident records from Shaanxi Province, June 2021 - August 2022. \\ \hline
    \end{tabularx}
    \label{tab:dur}
\end{table*}

\section{Road Accidents Statistical Analysis and Modeling}
\label{sec:analysis}
This collection of research advances the field of traffic safety through analyzing accident data using statistical and machine learning methods to enhance accident prediction, risk assessment, and emergency response \cite{ryder2019spatial, roland2021modeling, mohanty2023development, rashidi2022modeling, thapa2022overcoming, zhou2022urban, dias2023prediction, cai2023different, hou2023urban, thapa2024advancing}. The studies in this section use interpretable models to reveal the impact of various factors on predicting traffic accident risk and severity.  

\cite{dias2023prediction} addresses the gap in computational tools for analyzing road accident data by employing various regression algorithms on data from the Portuguese National Guard. Their study identified key variables correlated with accidents and demonstrated that the best neural network model achieved notable accuracy. However, the research was constrained by limited geographic scope and the exclusion of potential influential factors beyond the accident data itself, highlighting the need for a more comprehensive approach to understanding road safety dynamics.

Thapa et al. \cite{thapa2022overcoming} develop a novel approach to traffic crash prediction by discretizing inter-crash durations into 1-hour epochs with 15-minute intervals, allowing for a more detailed time analysis. They expanded crash data into these intervals and used a multinomial logit model with latent propensity functions. Their method showed that using a 25\% epoch-level sample could significantly reduce training time without greatly compromising accuracy. However, the study did not address unobserved heterogeneity across segments or the variability in traffic flow and geometry, limiting its real-world applicability. In a subsequent study, Thapa et al. \cite{thapa2024advancing} improve proactive crash prediction models by introducing a two-level nested logit model that predicts both crash occurrences and severities. They reformulate crash data into forecasting epochs and associated dynamic covariates such as traffic flow and speed. Their findings highlighted a 15\% sample size at the epoch level as optimal for balancing model performance and computational efficiency. The model showed high specificity but low sensitivity in crash prediction, suggesting room for improvement. The study's limitations include the assumption of a constant nesting parameter and the absence of considerations for time effects and unobserved heterogeneity within the panel data.

Mohanty et al. \cite{mohanty2023development} undertake a study to explore the differential roles of vehicles in crashes as perpetrators and victims, using 1,559 crash records from Visakhapatnam, India. The research applied Goodman-Kruskal tests and binary logistic regression to develop prediction models, alongside an ANN model. Key findings include the higher fatality rates for two-wheelers and heavy vehicles as perpetrators, and for pedestrians and self-hitting vehicles as victims. The logistic model achieved 75\% accuracy, improved to 81\% by adjusting the probability cut-off, while the ANN showed 75\% accuracy overall, reaching near 100\% for certain vehicle pairs. The main limitation lies in its geographical and data scope, potentially affecting the generalizability and accuracy of the findings for diverse regions and vehicle pairs. Dias et al. \cite{dias2023prediction} address the gap in computational tools for analyzing road accident data by leveraging records from the Portuguese National Guard database (2019-2021). The study applied various regression algorithms to identify patterns and variables influencing accidents, such as time intervals, weather, and day of the week, across different road types. The analysis highlighted the highest accident rates during the evening rush hour and an increase in accidents during rainy conditions, particularly on Fridays. The best-performing neural network model achieved a 0.49 mean absolute error (MAE) and 88\% accuracy on combined data, with slightly lower performance on motorways. However, the research was constrained by limited geographic scope and the absence of human mobility data, focusing solely on accidents without considering broader contributing factors.

A summary of studies in the category of traffic accidents analysis is provided in Table~\ref{tab:analysis}. 
\begin{table*}[h!]
    \centering
    \footnotesize
    \caption{Summary of Studies on Traffic Accidents Analysis Using Statistical and Machine Learning Methods}
    \begin{tabularx}{\linewidth}{|>{\hsize=.4\hsize}X|c|>{\hsize=.4\hsize}X|>{\hsize=1.8\hsize}X|>{\hsize=1.4\hsize}X|}
        \hline
        \rowcolor{lightblue} 
        \textbf{Study} & \textbf{Year} & \textbf{Main} \vfill \textbf{Approach} & \textbf{High-level Summary} & \textbf{Data Source} \\ \hline
        Ryder et al. \cite{ryder2019spatial} & 2019 & Statistical Modeling & This study proves a proportional relationship between Critical Driving Events and traffic accident frequency using spatial regression analysis. & Swiss Federal Roads Office's average daily traffic dataset and traffic accident dataset spanning 2011-2016. \\ \hline
        Roland et al. \cite{roland2021modeling} & 2021 & Macchine Learning & Proposed a MLP model to predict daily vehicular accident hotspots in Chattanooga, TN, enhancing emergency response. & Data from Hamilton County Emergency Services, Darksky API for weather, ETRIMS and ArcGIS for roadway info. \\ \hline
        Mohanty et al. \cite{mohanty2023development} & 2022 & Machine Learning & Study develops models using binary logistic regression and ANN to predict road crash fatalities based on vehicle roles. & Data from Visakhapatnam, India, 1559 road crash data from 2016-2017. \\ \hline
        Rashidi et al. \cite{rashidi2022modeling} & 2022 & Statistical Modeling & This study uses Holt-Winters forecasting to analyze how macroeconomic and traffic indicators affect road crash forecasting accuracy in Iran. & Data from 93-month Iranian provincial road traffic crash data from Mar-2011 to Dec-2018, sourced from Iranian traffic police. \\ \hline
        Thapa et al. \cite{thapa2022overcoming} & 2022 & Statistical Modeling & Developed a more efficient real-time crash prediction model using dynamic covariates and advanced sampling techniques. & Study uses crash data from Memphis, Tennessee's I-40 and I-55 using eTRIMs and Radar Detection System \\ \hline
        Zhou et al. \cite{zhou2022urban} & 2022 & Machine Learning & Developed a comprehensive framework using deep learning for accurate urban traffic accident prediction and safety improvement. & Real-vehicle test data (1022 sets, 23 experimenters, 6 routes) and open-source data (traffic volume, street view, road design). \\ \hline
        Dias et al. \cite{dias2023prediction} & 2023 & Machine Learning & Developed a tool to predict road accident risk using data mining algorithms on Portuguese National Guard data. & Portuguese National Guard database, 2019-2021, Setubal, Portugal.  \\ \hline
        Cai and Di \cite{cai2023different} & 2023 & Machine Learning & Developed a time series-based count data model for daily traffic crash prediction using machine learning and ARIMA. & Traffic accident data from a freeway in China, including weather and traffic flow. \\ \hline
        Hou \cite{hou2023urban} & 2023 & Machine Learning & Developed predictive models using Random Forest and Xgboost for analyzing factors affecting road traffic accident severity. & The data source is unclear.  \\ \hline
        Thapa et al. \cite{thapa2024advancing} & 2024 & Statistical Modeling & Introduced a novel framework for predicting crashes and their severities using a duration-based model. & 2019 crash data from ETRIMS and traffic data from RDS in Memphis and Chattanooga, Tennessee, totaling 2,375 crashes. \\ \hline
    \end{tabularx}
    \label{tab:analysis}
\end{table*}

\section{Future Directions of Research}
\label{sec:future}
In this section, we outline potential directions for future research in the field of traffic accident analysis, modeling, and prediction. These suggestions are derived from an extensive examination of current studies and the challenges they present, with the goal of identifying opportunities for advancement. Although we aim to provide a wide range of recommendations, it is essential to understand that these do not cover every possible area of research but are meant to encourage further exploration. The following points detail specific areas of interest highlighted in our review:

\begin{itemize}
    \item \textbf{Scalable and Diverse Data for Accident Prediction:} Current studies on accident analysis and prediction often focus on data with a limited number of variables. However, expanding the range of input variables can have a significant impact, making the models more useful for real-world applications. An interesting direction for future research could be to include a more comprehensive set of datasets, such as information related to construction, population density, and real-time traffic conditions. This approach aims to improve the accuracy of high-resolution accident prediction models by ensuring the data reflects the complexity of urban environments, thereby enhancing the models' applicability to various cities. It is important to note that access to such multi-modal data may be limited, particularly at a large scale. Researchers may start by exploring data from smaller areas to demonstrate a proof of concept, and then expand their work by collaborating with official and state traffic authorities and agencies. This collaboration could provide researchers with access to the necessary data to develop more robust and comprehensive accident prediction models that can be applied across different urban settings.

    \item \textbf{Interpretable Models for Traffic Management:} While employing complex machine learning models can lead to improved performance, it is crucial to have an interpretable model, even if it is only to some extent. This aspect is important for enhancing the adaptability of the model in real-world applications. In other words, if researchers expect their work to be employed outside the scope of academic research, they must invest in areas that make their model and the corresponding inferences understandable for human stakeholders with little or no machine learning domain knowledge. Therefore, the suggestion is to develop interpretable machine learning models that can be integrated into intelligent transportation systems for active traffic management. This involves exploring explainable AI techniques to clarify model decisions, thereby enabling stakeholders to make informed decisions based on the models' predictions.
    
    \item \textbf{Advanced Machine Learning Techniques for Enhanced Prediction:} Despite significant advances in machine learning and artificial intelligence, most contemporary studies still rely on relatively simple machine or deep learning methods. This presents an opportunity to integrate more sophisticated models, particularly for modeling multi-modal data and avoiding underfitting by better capturing underlying data relationships. Such integration can optimize input variable utility and yield more precise real-time predictions. The goal should be to incorporate state-of-the-art machine learning and deep learning algorithms to enhance both prediction accuracy and efficiency. This endeavor would encompass the exploration of transfer learning (to capitalize on models and techniques originating from domains beyond accident modeling), transformer models (to more effectively address long-range dependencies within the data), graph neural networks (to more accurately represent accidents within intricate road networks and their interactions with diverse environmental factors), self-supervised learning (to uncover latent phenomena without the need for extensive human supervision and at a relatively low cost), among other pertinent techniques. Such an approach aims to navigate the complexity inherent in traffic data and the task of accident prediction, thereby contributing to the advancement of this critical field. 

    \item \textbf{Integration into Autonomous Vehicles and Advanced Driver Assistance Systems (ADAS):} As an emerging field of significant interest in both research and industrial development, a portion of the studies discussed in this manuscript is directly relevant to autonomous vehicle technology. However, it should be noted that the majority of these studies may not be seamlessly applicable to such applications. Particularly, those relying on computational methods and general data may face transferability issues. Given the significant advancements in autonomous driving technologies and ADAS, there is an increasing demand for developing models that can be effectively utilized in these contexts to yield precise outcomes, characterized by high rates of true positive detections and minimal false alarms. It is crucial to acknowledge that conducting research in this domain necessitates access to specific types of data, such as those obtained from cameras, radar, and LiDAR sensors, on a large scale. Unfortunately, such data is not widely available to the public, contributing to the scarcity of pertinent research in this field. Additionally, investigations in this area demand substantial computational resources, which may further restrict their broad adoption. Despite these challenges, there exists a significant void in the existing literature regarding the development of systems suitable for application in autonomous driving and ADAS. This gap presents a substantial opportunity for research and development efforts aimed at enhancing safety measures and transforming the current state-of-the-art in vehicle automation.
    
    \item \textbf{Integration of Real-time Alert Systems:} The practical application of accident analysis and forecasting is most effective when combined with real-time systems that continuously monitor traffic conditions and various other inputs to provide accurate safety predictions. The objective is to validate and incorporate these predictive models into real-time alert systems and traffic management infrastructures. This involves testing the models in actual traffic environments to confirm their capacity to lower accident occurrences and enhance overall road safety. Furthermore, it necessitates access to diverse and realistic input data, models trained on real-world scenarios, and a proven history of accurate predictions in practical settings. However, according to the research examined in this study, there is a notable scarcity of studies that meet all these criteria, highlighting a significant area for future research and development in this field.
    
    \item \textbf{Tailoring Models to Various Environmental Factors:} Although many studies have designed their methodologies to utilize a range of inputs, such as traffic, weather, and road network data, there is a general shortfall in validating these models across different environmental conditions. This underscores the necessity for rigorous testing and customization of predictive models to suit a variety of road surfaces, weather patterns, and driving contexts, ensuring their dependability and effectiveness. This process should involve the integration of more comprehensive road layout information, specific weather conditions, and up-to-date traffic data. Moreover, creating these models will likely require access to large datasets that span extensive time frames to effectively capture a wide range of environmental and situational variables. 
    
    \item \textbf{Enhancing Data Acquisition and Handling:} A primary obstacle in the realm of traffic accident analysis and forecasting pertains to the collection and utilization of data. To address this issue, various studies have devised their own methods for either gathering data anew or obtaining it from formal entities, such as traffic management bodies. Frequently, the data employed in these studies suffer from limitations in accuracy or comprehensiveness, leading to a diminished applicability of the proposed solutions across different contexts. Although efforts have been made to bridge this data gap to a certain degree, particularly in nations like the United States and the United Kingdom, a substantial void remains. Overcoming this challenge is unlikely to be feasible through academic research endeavors alone, due to the typically constrained resources. A broader, coordinated effort is essential to gather pertinent data—encompassing traffic patterns, accident occurrences, and meteorological conditions—and to subsequently make this information widely accessible to the global research community. Such an initiative would significantly aid in the creation of more precise and inclusive predictive models. However, it's important to recognize that this endeavor demands substantial collaborative engagement across multiple sectors, involving academic institutions and traffic regulatory agencies, and would likely necessitate a non-profit framework. Furthermore, it is important to acknowledge that a universally effective model may be an unrealistic expectation, given the diverse nature of regions worldwide. Consequently, acquiring high-quality, region-specific data emerges as a critical prerequisite for the formulation of adaptable and broadly applicable traffic accident prediction models. 
    
    \item \textbf{Generalization and Adaptability of Models:} As previously noted, it is unrealistic to anticipate that a singular predictive model will perform equally well across all geographical areas. Therefore, it is imperative to investigate the adaptability of these models to varied geographic locales and traffic scenarios. This entails the inclusion of diverse driving data and the evaluation of these models using datasets from distinct cities and territories. Concurrently, there exists a significant potential for employing methods such as transfer learning. This approach allows for the utilization of insights gained from one region in the development or refinement of a model for another, potentially novel, region.
    
    \item \textbf{Broader Implications for Traffic Safety and Management:} Although the primary emphasis of this manuscript was on predicting traffic accidents and analyzing their severity, it is pertinent to acknowledge other critical facets of traffic safety and management that were not covered in detail within this study. In this context, promising research directions might involve the formulation and application of predictive models to a wider array of traffic safety and management issues. These could include the identification of locations prone to severe accidents, pinpointing hazardous areas, studying spatiotemporal changes in traffic accidents over time for specific locations such as cities or roads, and monitoring traffic conditions in real-time. The integration of such models with current traffic management infrastructures could substantially augment their effectiveness and contribute to enhancing overall road safety. 
    
\end{itemize}
\section{Summary and Conclusions}
\label{sec:conc}
Traffic accidents remain a severe global public health crisis according to the latest reports from the WHO. Given the significance of this issue, numerous research studies over the past decade have focused on addressing traffic accidents by applying machine learning or statistical analysis techniques. These techniques aim to either predict various aspects of accidents or uncover potential root causes to better prevent them in the future. This study provides a comprehensive review of related research in the domain, specifically focusing on studies from the past five years that could potentially define the state-of-the-art. Our review surveys five categories of research: accident risk (or occurrence) prediction, frequency prediction, severity prediction, duration prediction, and general accident data analysis. We reviewed over 190 recent studies from leading journals and conferences in the field.

In reviewing each paper, we focused on the methodology, data sources, results, and main limitations to inspire potential future research directions. In addition to the comprehensive literature review, we discuss several aspects that present opportunities for future research. For instance, the research community may consider utilizing more diverse data sources to build and validate their proposals, enhancing reliability and reproducibility. Another potential area is better integration and more research concerning autonomous driving, given the emerging industry direction and the need for fundamental research in this domain. Integrating research with real-time systems and employing various environmental factors could also be aspects that strengthen proposals in this area.

Overall, this survey paper sought to provide a comprehensive overview of the state-of-the-art, highlight limitations of various studies, and lay out a clear picture for future research in the field of traffic accident analysis and prediction using machine learning and statistical techniques. By synthesizing recent advancements and identifying gaps, this review aims to guide further investigations towards significantly reducing traffic-related deaths and injuries, aligning with the WHO targets.

\bibliography{main}

\newpage
\clearpage

\begin{appendices}
\section{List of Acronyms}
\label{sec:appendix}
\setlength{\extrarowheight}{2pt}

\begin{table}[ht]
\centering
\caption{List of Acronyms and Descriptions}
\begin{tabular}{ll | ll}
\toprule
\textbf{Acronym} & \textbf{Description} & \textbf{Acronym} & \textbf{Description} \\
\midrule
ADAS & Advanced Driver-Assistance Systems & ANN & Artificial Neural Network \\
AUC & Area Under Curve & AUROC & \begin{tabular}{@{}l@{}} Area Under the Receiver \\ Operating Characteristics\end{tabular} \\
AutoML & Automated Machine Learning & BART & Bayesian Additive Regression Trees \\
BERT & \begin{tabular}{@{}l@{}} Bidirectional Encoder \\ Representations from Transformers\end{tabular} & BR & Bayesian Regularization \\
BRF & Balanced RF & CART & Classification and Regression Trees \\
CNN & Convolutional Neural Network & DCGAN & \begin{tabular}{@{}l@{}} Deep Convolutional Generative \\ Adversarial Network\end{tabular} \\
DCGANs & \begin{tabular}{@{}l@{}} Deep Convolutional Generative \\ Adversarial Networks\end{tabular} & DJ & Decision Jungle \\
DL & Deep Learning & DNN & Deep Neural Network \\
DNNs & Deep Neural Networks & DOT & Department of Transportation \\
DT & Decision Tree & EVT & Extreme Value Theory \\
g-force & Gravitational Force Equivalent & GA & Genetic Algorithms \\
GANs & Generative Adversarial Networks & GBM & Gradient Boosting Machine \\
GCNs & Graph Convolutional Networks & GMM & Gaussian Mixture Model \\
GPR & Gaussian Process Regression & GPS & Global Positioning System \\
GRNN & General Regression Neural Network & GRU & Gated Recurrent Unit \\
HSM & Highway Safety Manual & KLT & Kanade-Lucas-Tomasi \\
LiDAR & Light Detection and Ranging & LSTM & Long Short-term Memory \\
MAE & Mean Absolute Error & MAPE & Mean Absolute Percentage Error \\
ML & Machine Learning & MLP & Multi-Layer Perceptron \\
NLP & Natural Language Processing & PART & Partial Decision Tree \\
PCA & Principal Component Analysis & PNN & Probabilistic Neural Network \\
PSO & Particle Swarm Optimization & R-CNN & Region-based Convolutional Neural Network \\
RBM & Restricted Boltzmann Machine & RENB & Random Effects Negative Binomial \\
RF & Random Forest & RNN & Recurrent Neural Network \\
SMOTE & Synthetic Minority Over-sampling Technique & SSAE & Stacked Sparse Autoencoder \\
SVM & Support Vector Machine & SVR & Support Vector Regression \\
TF-IDF & Term Frequency-Inverse Document Frequency & TMC & Traffic Message Channel \\
UAV & Unmanned Aerial Vehicle & VANET & Vehicular Ad-hoc NETwork \\
WGAN & Wasserstein Generative Adversarial Network & WHO & World Health Organization \\
XGBoost & eXtreme Gradient Boosting & & \\
\bottomrule
\end{tabular}
\label{tab:acronyms}
\end{table}

\end{appendices}

\end{document}